%% file: acl_latex.tex
\pdfoutput=1

\documentclass[11pt]{article}

\usepackage[preprint]{acl}
\usepackage{float}
\usepackage{times}
\usepackage{latexsym}
\usepackage{comment}
\usepackage[most]{tcolorbox}
\usepackage{xcolor}

\usepackage[T1]{fontenc}

\usepackage[utf8]{inputenc}

\usepackage{microtype}

\usepackage{inconsolata}

\usepackage{graphicx}

\usepackage{hyperref}
\usepackage{url}
\usepackage{graphicx}
\usepackage{tabularx}
\usepackage{array}
\usepackage{algorithm}
\usepackage{algpseudocode}
\usepackage{amsmath}

\title{Concept Distillation from a Strong Model \\ to Weak Models via Hypotheses-to-Theories \\ Prompting}

\title{Concept Distillation from Strong to Weak Models via Hypotheses-to-Theories Prompting}

\author{
    Emmanuel Aboah Boateng,
    Cassiano O. Becker, Nabiha Asghar, Kabir Walia\\ 
    \textbf{Ashwin Srinivasan, Ehi Nosakhare,  Soundar Srinivasan, Victor Dibia} \\
    Microsoft \\
    \texttt{\{emmanuelab, casbecker, kabirwalia, ashwinsr, ehinosakh,} \\
    \texttt{sosrini, victordibia\}@microsoft.com}
}

\begin{document}
\maketitle
\begin{abstract}
    \input{0abstract_v1}

\end{abstract}

\section{Introduction}
\label{sec: intro}
\input{1introduction_v1}

\section{Related Work}
\label{sec: related_work}
\input{6related_work_v1}

\section{Background}
\label{sec: background}
\input{2background_v1}

\section{Concept Distillation Framework}
\label{sec: concept_distillation}
\input{3concept_distillation_v1}

\section{Experiments}
\label{sec: expts}
\input{4experiments_v1}

\section{Results and Analyses}
\label{sec: results}
\input{5results_v1}

\section{Conclusion}
\label{sec: conclusion}
\input{7conclusion_v1}

\bibliography{references}

\appendix


\input{appendix}

\end{document}

%% file: 0abstract_v1.tex
Hand-crafting high quality prompts to optimize the performance of language models is a complicated and labor-intensive process. Furthermore, when migrating to newer, smaller, or weaker models (possibly due to latency or cost gains), prompts need to be updated to re-optimize the task performance. We propose \textit{Concept Distillation} (CD), an automatic prompt optimization technique for enhancing weaker models on complex tasks. CD involves: (1) collecting mistakes made by weak models with a base prompt (initialization), (2) using a strong model to generate reasons for these mistakes and create rules/concepts for weak models (induction), and (3) filtering these rules based on validation set performance and integrating them into the base prompt (deduction/verification). We evaluated CD on NL2Code and mathematical reasoning tasks, observing significant performance boosts for small and weaker language models. Notably, Mistral-7B's accuracy on Multi-Arith increased by 20\%, and Phi-3-mini-3.8B's accuracy on HumanEval rose by 34\%. Compared to other automated methods, CD offers an effective, cost-efficient strategy for improving weak models' performance on complex tasks and enables seamless workload migration across different language models without compromising performance.

%% file: 1introduction_v1.tex
Large language models (LLMs) have shown remarkable capabilities for various downstream tasks. An inexpensive alternative to training and fine-tuning, \emph{prompt engineering} has emerged as a powerful method to control and optimize the outputs from LLMs. Prompt engineering is enabled by the in-context learning (ICL) capability of LLMs \cite{dong2022survey}, which allows us to apply LLMs to new tasks by providing them with a suitable input prompt that contains relevant information and instructions \cite{xie2021explanation}.

\begin{figure}[t]
    \centering
    \includegraphics[width=\columnwidth]{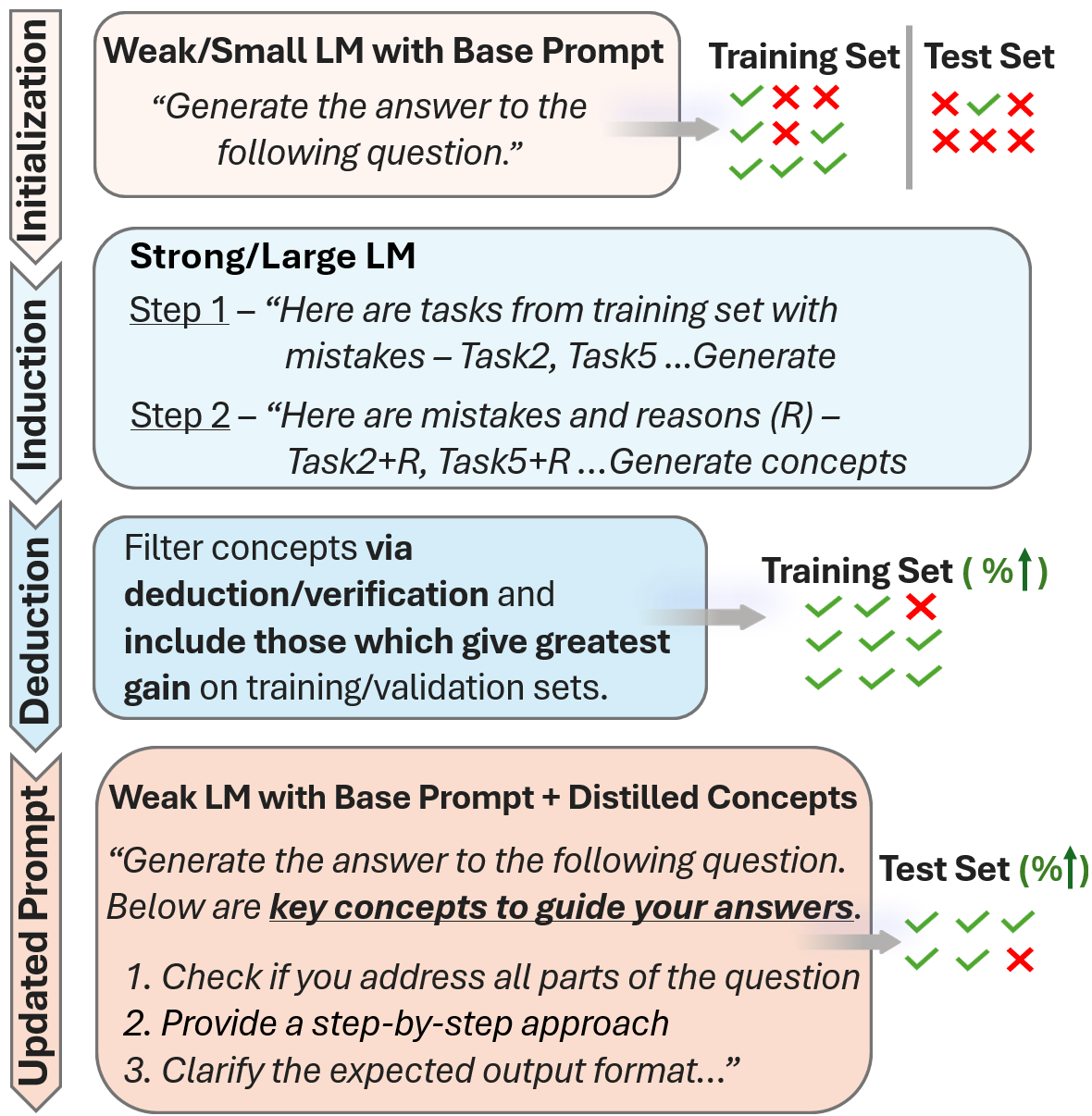}
    \caption{High-level illustration of concept distillation for prompt construction.}
    \label{fig:highlight}
\end{figure}

As such, crafting high-quality prompts can be a challenging and labor-intensive process. Finding the right instructions can require several rounds of trial-and-error experimentation. Further, the same prompt may not work for different tasks, models, or domains \cite{lu2023prompts, rubin2021learning}. Importantly, weak models such as GPT-3.5 or Mistral-7B often lack the same reasoning capabilities as strong models such as GPT-4o, and as a result, struggle with complex and high-reasoning tasks \cite{edwards2024language,liang2023encouraging}. This leads to significant performance gaps between stronger and weaker models for such tasks. Conversely, practical reasons (e.g., lower runtime latency, cost, and memory footprint) may still motivate and impose the use of weak models in practical applications \cite{xia2023flash, hadi2023survey}. 
While fine-tuning methods such as LoRA~\cite{hu2022lora} may close this gap, they involve modifying the model’s parameters --- thus making it task-specific and limiting its reuse across different contexts. In particular, these approaches require fine tuning infrastructure and know-how, which may not be available or accessible in many practical scenarios. In contrast, our CD approach preserves the model’s parameters, allowing the model to remain flexible for various tasks, and requiring only prompt-engineering level of access.

Another key area that the current work addresses is the efficient adaptation of prompts for various models. A primary challenge is transitioning prompts from an existing model, such as GPT-4, to a newly released variant like GPT-4o. It is essential to recognize that different models may respond uniquely to the same prompts. As such, there is the need for strategies that effectively modify and tailor existing prompts to maintain alignment with new or evolving models.

In this paper, we introduce \textit{concept distillation} (CD), an automated prompt optimization technique. CD improves the performance of weak/small language models on complex tasks by distilling key rules, concepts, or examples from a strong/large model via hypotheses-to-theories prompting. These distilled concepts are then verified and used to guide a weak model, enabling it to produce more accurate responses, all without the need for fine-tuning. The structured approach within the CD framework ensures that these distilled concepts are sufficiently general to be transferable across various other language models. 
\mbox{Figure~\ref{fig:highlight}} shows a high-level illustration of the concept distillation for prompt optimization approach. Overall, this paper makes the following contributions:
\begin{itemize}
    \item We introduce the notion of \textit{concept distillation}, in which a strong model is used to derive new concepts (i.e.,  specific prompt instructions) to help a weak model improve its performance on complex tasks, thereby enabling greater adaptability of the weak model in various applications (see Fig.~\ref{fig:knowledge_vs_concept}).
    \item Building on time-tested principles of scientific discovery, we propose the \textit{hypotheses-to-theories} prompt optimization framework, which leverages the strong model's ability to perform inductive and deductive reasoning over the weak model's deficiencies (see Sections \ref{sec: background} and \ref{sec: concept_distillation}).
    \item We demonstrate that the prompt optimization framework enables efficient adaptation of prompts across different language models (LMs). The distilled concepts are transferable, allowing for quick and effective updates in response to new model releases or changes, ensuring continued optimal performance (see Section \ref{subsec: concepts_transfer}).
    \item We perform a systematic experimental evaluation on different tasks (NL2Code: HumanEval, Mathematical Reasoning: GSM8K/Multi-Arith) with various weak models (GPT-3.5 Turbo, Claude 2.1, Phi-3-mini-3.8B, Mixtral-8x7B, and Mistral-7B), and show that the proposed approach significantly reduces the performance gap between the weak and strong models (see Sections \ref{sec: expts} and \ref{sec: results}).
\end{itemize}

%% file: 6related_work_v1.tex
Given the significance and broad-scale effectiveness of prompt engineering, there have been several efforts to perform automated prompt optimization and generation. 
These methods typically involve an iterative algorithm consisting of several steps -  an initially generated prompt, scoring of the prompt, and regeneration of the prompt using the score as an improvement signal, till a stopping criteria is met \cite{zhou2022large_ape, hu2023evoke, pryzant2023automatic_gradient, ye2023prompt, wang2023promptagent, deng2023rephrase, guo2023connecting_evoprompt}. We propose an approach that augments this framework for prompt optimization through the distillation of \textit{concepts}, and introduces an explicit verification step to demonstrate relative performance improvements for a small model. 

Our method is inspired by several recent works. APE \cite{zhou2022large_ape} deduces an initial prompt from training samples, and then uses an LLM to refine and generate new semantically similar prompt candidates. However, prompts are simply paraphrased during the refinement process, which is akin to random search in the prompt space. Evoke \cite{hu2023evoke} uses the same LLM to review and score the quality of a prompt, as well as to refine the prompt based on the reviewer feedback. \cite{zhu2023large} first uses an LLM to induce a rule library from a set of training examples, which are later sampled for dynamic prompt construction. This is followed by a deduction phase where these rules are evaluated based on their coverage and confidence. \cite{zhang2024context} generates 
 high and low-level concepts from mistakes using an LLM, and later uses the same LLM for solving tasks. There is no deduction phase to filter out the generated concepts. PE2 \cite{ye2023prompt} explores meta-prompt variants to guide LLMs to perform automatic prompt engineering. They introduce 3 meta-prompt components - two-step task description, context specification and step-by-step reasoning template to improve task performance.

In contrast to all these works, our method focuses on transferring capability from a large/strong model to a small/weak one by inducing \textit{concepts} mainly from the mistakes made on a task by the weak model. Additionally, our deduction phase filters out the generated concepts in a metric-driven way, which is a crucial additional step in our framework that improves task adaptability and performance of weak models.

Many other works explore various fundamentally different frameworks for automatic prompt optimization, and are noteworthy to mention here. There are text-based error-propagation techniques such as PromptAgent \cite{wang2023promptagent} which uses Monte Carlo Tree Search, and ProTeGi \cite{pryzant2023automatic} which mirrors the steps of gradient descent-like updates for prompts. TRAN \cite{yang2023failures} takes a different approach by accumulating failure-driven rules at inference time, enabling LLMs to iteratively improve without fine-tuning. Another category of related works employs parametric (non-interpretable) prompt optimization techniques, as opposed to edit-based ones \cite{su-etal-2022-transferability,zhong2024panda, wen2024hard}.

%% file: 2background_v1.tex
In this section, we explore the foundational concepts and terminologies central to this paper. This technique draws inspiration from human cognitive processes \cite{hunt2003concept, cherukunnath2022exploring}, particularly in how we acquire, refine, and apply knowledge and concepts across various domains. 

\begin{figure*}[thp]
    \centering
    \includegraphics[width=\textwidth]{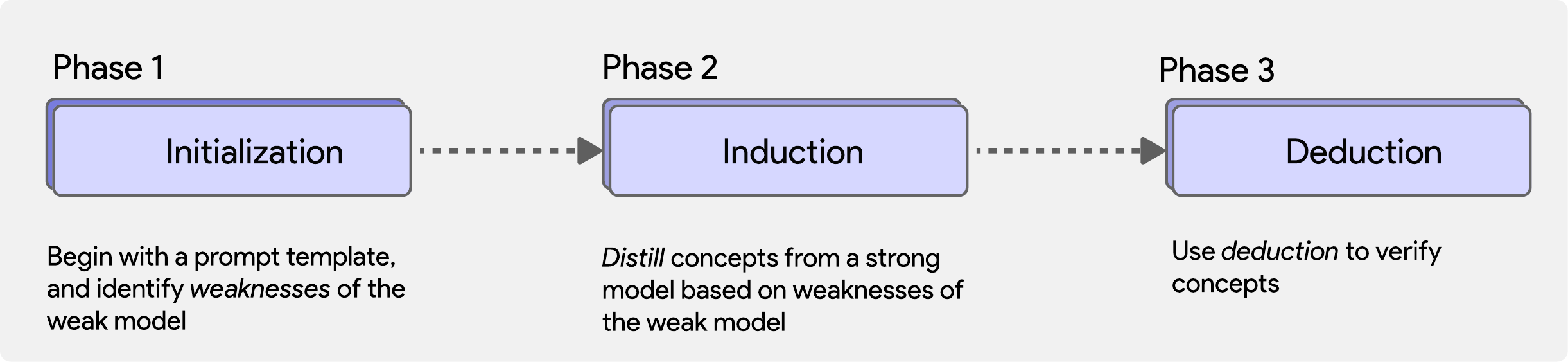}
    \caption{Workflow of concept distillation for prompt optimization.}
    \label{fig:intro_CD}
\end{figure*}

\textbf{Concept Distillation: distinction from Knowledge Distillation} depicted in Fig.~\ref{fig:intro_CD}. The core of our technique is encapsulated in the process of `concept distillation'. This process involves the transfer of concepts from a stronger LM (referred to as the `teacher') to a weaker LM (referred to as the `student'). The differentiation between knowledge and concept distillation is critical. Unlike traditional knowledge distillation \cite{phuong2019towards}, which focuses on the explicit transfer/update of learned weights and biases through intensive training or fine-tuning procedures, concept distillation emphasizes the induction of general concepts, rules, examples, or key ideas from the teacher model, applying them to the student model solely via in-context learning (ICL), without necessitating extensive training or fine-tuning. Figure \ref{fig:knowledge_vs_concept} depicts the distinction between knowledge and concept distillation.

\textbf{Hypotheses, Theories, and Reasoning: frameworks for conceptual transfer}.
Our approach is deeply rooted in the scientific methodologies of hypothesis generation, experimental validation, and theory  \cite{scerbo2019research}. A hypothesis, in this context, is a proposition based on limited evidence, serving as a foundation for further investigation that could culminate in a theory, i.e., a well-substantiated explanation of a phenomenon. This framework is critical in concept distillation, where hypotheses derived from observations are validated through experimental evidence to form theories that explain the underlying principles or phenomena.

The transformation from hypotheses to theories is facilitated by two modes of reasoning: \textit{inductive} and \textit{deductive} reasoning. Inductive reasoning involves deriving general rules from specific observed facts, whereas deductive reasoning entails deriving new facts from established facts and rules. Deductive reasoning allows us to apply general principles to specific cases to derive accurate conclusions. These modes of reasoning allow the extrapolation of concepts from inductive reasoning and the application of these concepts to new, unseen instances.

Drawing parallels to the human process of scientific discovery \cite{bradford2022science}, our technique mirrors the iterative cycle of observation, hypothesis formulation, experimentation, and theory development. This analogy highlights the integration of inductive and deductive reasoning in forming robust concepts that not only explain observed phenomena but also predict outcomes in unseen scenarios.


%% file: 3concept_distillation_v1.tex

Our technique consists of three main phases: \textit{initialization}, \textit{induction}, and \textit{deduction from verification}. 

\textbf{Initialization phase}. Phase 1 starts with a base prompt template, which can be either an existing prompt we aim to modify (for a strong, large model we aim to replace), a generated prompt using an off-the-shelf algorithm, or one manually crafted by domain experts, serving as a foundation for subsequent refinement. In this phase, we assess the strengths and weaknesses (mistakes) of the weaker model regarding the intended task. The primary goal here is to pinpoint areas and examples where the weaker model struggles, enabling us to induce concepts that aid in reasoning in these specific areas. It is important to focus on the model's weaknesses, avoiding unnecessary adjustments in areas where the model already performs well. 

\textbf{Induction phase}. Phase 2 involves the induction of concepts from a strong model, such as GPT-4, tailored to address the identified weaknesses and mistakes of the weaker model. The aim is to enhance the weaker model's performance by equipping it with these newly induced concepts. During this process, we use the strong model to reason through the facts or questions presented to the weak model, the incorrect responses it generated, and the correct answers, in order to generate general concepts that can overcome the mistakes of the weak model.

\textbf{Deduction from verification phase}. Phase 3 is the deduction-from-verification process. The assumption is that not all induced rules/concepts or examples qualify as useful distilled concepts. This phase uses a deduction process to verify the induced rules and examples. Rules that qualify as having broad coverage and high prediction confidence are accepted as distilled concepts. Consequently, they are added to the initial prompt template that we started with to form an improved, updated prompt. After adding the induced concepts to the base prompt template, a verification process is applied to filter the concepts. Either the strong model can be used to generate test examples similar to the weaknesses identified earlier or similar examples from a validation set can be used for the verification. The weaker model is required to accurately address all validation examples with a level of certainty or probability that meets or exceeds a specific predefined threshold before we accept the induced concepts as distilled concepts and integrate them into its prompt. This ensures that the final prompt effectively addresses the weak model's shortcomings, leading to improved performance. 

\begin{figure*}[thp]
    \centering
    \includegraphics[width=\textwidth]{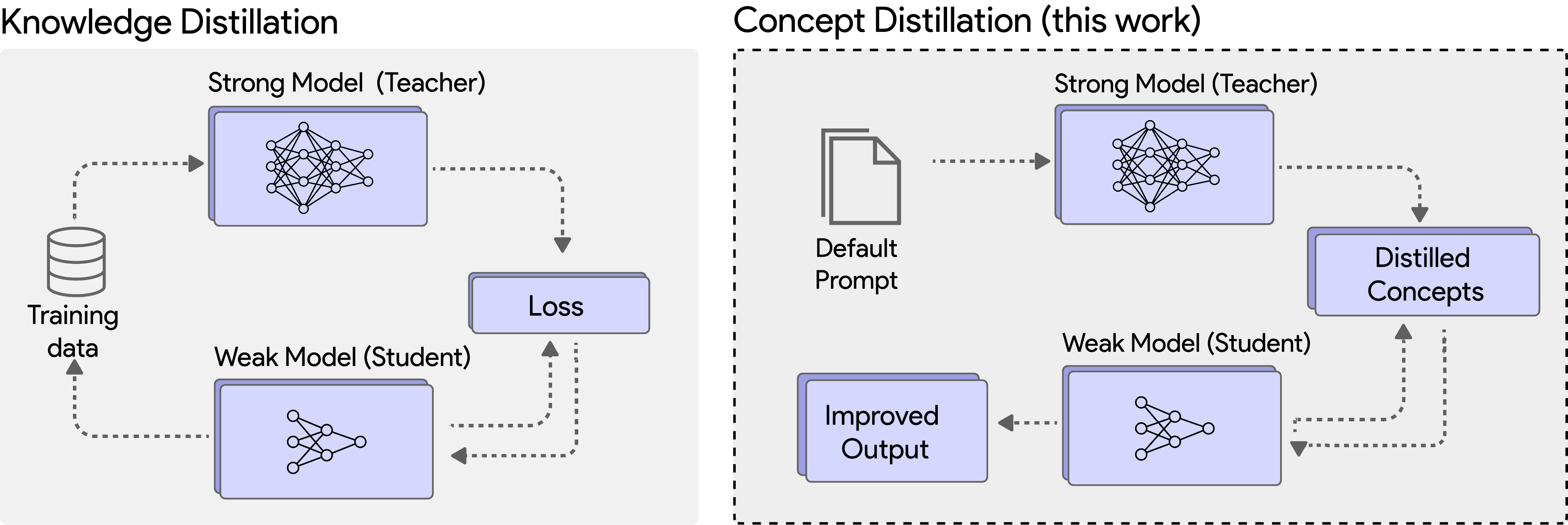}
    \caption{Distinction between knowledge and concept distillation.}
    \label{fig:knowledge_vs_concept}
\end{figure*}

Algorithm \ref{alg:one} succinctly captures the proposed CD framework. It details the three key phases—initialization (see Fig. \ref{fig:cd_initialization}), induction (see Fig. \ref{fig:cd_induction}), and deduction/verification (see Fig. \ref{fig:cd_deduction}). 
The definitions and descriptions of the notations and processes as well as the prompts used in the algorithm are provided in Appendix \ref{sec: Notations}.
A detailed description of the concept distillation process with a walk-through example is provided in Appendix \ref{sec: hypothetical_example}.

\begin{algorithm}
\caption{Hypotheses-to-Theories CD} \label{alg:one}
\begin{algorithmic}[1]
\Require Strong model \(M_s\), Weak model \(M_w\), Training set \(D\), Initial prompt \(p_0\)
\Statex \texttt{(i) Initialization:}
\State \(C \gets \emptyset\) \Comment{Set of distilled concepts}
\State \(p \gets p_0\) \Comment{Initialize prompt}
\For{each $(x_i, y_i) \in D$}
    \State $y_{w} \gets M_w(x_i, p)$ 
    \If{$y_{w} \neq y_i$}
        \Statex \texttt{(ii) Induction:}
        \State $R \! \gets \! \text{InduceConcept}(M_s, x_i, y_i, y_{w}, p)$
        \State $C \! \gets \!  C \cup R$ 
    \EndIf
\EndFor

\Statex \texttt{(iii) Deduction --> Verification:}
\For{each concept $c \in C$}
    \State $\text{ValidateConcept}(M_s, M_w, c, D)$ 
\EndFor

\Statex \texttt{Prompt Update:}
\State $p \gets \text{UpdatePrompt}(p_0, C)$ 
\State \Return $p$
\end{algorithmic}
\end{algorithm}

%% file: 4experiments_v1.tex
We focus on three benchmark datasets: NL2Code (HumanEval) \cite{chen2021evaluating_humaneval}, Multi-Arith \cite{roy-roth-2015-solving}, and GSM8K \cite{cobbe2021training}. HumanEval involves generating code from natural language prompts, while Multi-Arith and GSM8K evaluate arithmetic and mathematical reasoning, requiring step-by-step solutions.

We compare our approach with methods such as Automatic Prompt Engineering (APE) \cite{zhou2022large}, Iterative APE \cite{zhou2022large}, Chain of Thought (CoT) \cite{wei2022chain}, Prompt Engineering a Prompt Engineer (PE2) \cite{ye2023prompt}, and Automatic Prompt Optimization (APO) \cite{pryzant2023automatic}. We evaluate CD using GPT-3.5 Turbo, Claude 2.1, Phi-3-mini-3.8B, Mixtral-8x7B*, and Mistral-7B, with GPT-4o for concept distillation. Our experiments focus on improving weak models' performance through CD and testing the transferability of optimized prompts across models. We split each dataset into training and test sets for prompt optimization and evaluation, comparing our method with state-of-the-art techniques.

%% file: 5results_v1.tex
In Table \ref{tab:tab1} we summarize the performance of various models on the HumanEval test set, using only the base prompt and after CD (using the updated prompt with concepts). Notably, with base prompt alone, the strong model GPT-4o achieved a perfect score (100\%); in comparison, the weak models performed poorly. However, when using the updated prompt with concepts distilled using the CD technique, we observe significant performance boosts for the weak models.

\begin{table}[h]
    \normalsize
    \centering
    \begin{tabularx}{0.5\textwidth}{c|X|X}
    \hline
    \textbf{Model} & \textbf{Base prompt} & \textbf{CD} \\
    \hline
        GPT-3.5 & 0.85 & \textbf{0.96$_{(+11\%)}$} \\
    \hline
        Claude 2.1 & 0.89 & \textbf{0.99$_{(+10\%)}$}  \\
    \hline
        Phi-3-mini-3.8B & 0.48 & \textbf{0.82$_{(+34\%)}$} \\
    \hline
        Mixtral-8x7B* & 0.83 & \textbf{0.95$_{(+12\%)}$} \\
    \hline
        Mistral-7B & 0.89 & \textbf{0.96$_{(+7\%)}$} \\
    \hline
    \end{tabularx}
    \caption{Accuracy results on the HumanEval dataset for each model using both a base prompt and its optimized prompt based on CD. Corresponding results for the GSM8K dataset are presented in Appendix~\ref{sec: additional_GSM8K}}.
    \label{tab:tab1}
    \normalsize
\end{table}

\begin{table}[h]
\normalsize
    \centering
    \begin{tabularx}{0.5\textwidth}{c|X|X}
    \hline
    \textbf{Model} & \textbf{Base prompt} & \textbf{CD} \\
    \hline
        GPT-3.5 & 0.89 & \textbf{0.95$_{(+6\%)}$} \\
    \hline
        Claude 2.1 & 0.62 & \textbf{0.91$_{(+29\%)}$}  \\
    \hline
        Phi-3-mini-3.8B & 0.81 & \textbf{0.83$_{(+2\%)}$} \\
    \hline
        Mixtral-8x7B* & 0.72 & \textbf{0.88$_{(+16\%)}$} \\
    \hline
        Mistral-7B & 0.41 & \textbf{0.67$_{(+20\%)}$} \\
    \hline
    \end{tabularx}
    \caption{Accuracy results on the Multi-Arith dataset: Results are presented for each model using both a base prompt and its corresponding optimized prompt based on CD.}
    \label{tab:tab2}
\end{table}

\begin{table*}[h]
\normalsize
    \centering
    \begin{tabular}{|c|c|c|c|c|c|}
        \hline
        \multicolumn{1}{|c|}{} & \multicolumn{5}{c|}{\textbf{Model}} \\ \cline{2-6} 
        \textbf{Method} & \textbf{GPT-3.5} & \textbf{Claude-2.1} & \textbf{Phi-3-mini-3.8B} & \textbf{Mixtral-8x7B*} & \textbf{Mistral-7B} \\ \hline
        APE & 0.93 & 0.96 & 0.83 & 0.73 & 0.71 \\ \hline
        CoT & 0.45 & 0.82 & \textbf{0.91} & 0.88 & 0.87 \\ \hline
        \textbf{CD} & \textbf{0.96} & \textbf{0.99} & 0.82 & \textbf{0.95} & \textbf{0.96} \\ \hline
    \end{tabular}
    \caption{Accuracy comparison on the \textit{HumanEval} dataset between APE, CoT, and CD. Comparison with alternative methods based on specifically-built method implementations.}
    \label{tab:tab3}
\normalsize
\end{table*}

\begin{table*}[h]
\normalsize
    \centering
    \begin{tabular}{|c|c|c|c|c|c|}
        \hline
        \multicolumn{1}{|c|}{} & \multicolumn{5}{c|}{\textbf{Model}} \\ \cline{2-6} 
        \textbf{Method} & \textbf{GPT-3.5} & \textbf{Claude-2.1} & \textbf{Phi-3-mini-3.8B} & \textbf{Mixtral-8x7B*} & \textbf{Mistral-7B} \\ \hline
        APE & 0.63 & 0.43 & 0.78 & 0.84 & 0.65 \\ \hline
        CoT & 0.71 & 0.48 & 0.83 & 0.85 & \textbf{0.72} \\ \hline
        Iterative APE & 0.69 & 0.39 & 0.79 & 0.83 & 0.69 \\ \hline
        APO & 0.79 & 0.53 & 0.77 & 0.86 & 0.68 \\ \hline
        PE2 & 0.78 & 0.49 & 0.83 & 0.86 & 0.67  \\ \hline
        \textbf{CD} & \textbf{0.95} & \textbf{0.91} & \textbf{0.83} & \textbf{0.88} & 0.67 \\ \hline
    \end{tabular}
    \caption{Accuracy comparison on the \emph{Multi-Arith} dataset of different models and methods. Comparison with alternative methods based on optimized prompts as reported in \cite{ye2023prompt}.}
    \label{tab:tab4}
    \normalsize
\end{table*}

Firstly, we observe a performance increase of 11\% for the GPT-3.5 Turbo model, raising its accuracy from 0.85 to 0.96. Claude 2.1 nearly achieved a perfect score, improving from 0.89 to 0.99, an increase of 10\%, indicating that CD is effective in optimizing prompts even for models that initially perform well. The most notable performance gain was observed with the smallest model, Phi-3-mini-3.8B, which saw a substantial improvement of 32\%, from 0.48 to 0.82. Across all models evaluated, there was a significant performance increase compared to the base prompt evaluation, with an average performance increase of ~13\%. 

In Table \ref{tab:tab2} we summarize the results on the Multi-Arith dataset. We observe a 6\% performance gain for the GPT-3.5 Turbo model, a significantly larger gain for the Claude 2.1 model with a 29\% increase in accuracy from 0.62 to 0.91, and a similarly large 20\% accuracy gain for the Mistral-7B model. On average, weak models observed a performance lift of ~15\% on the Multi-Arith mathematical reasoning task.

The results in Table \ref{tab:tab1} and \ref{tab:tab2} provide evidence that Concept Distillation enhances the capabilities of weaker and smaller models, helping them overcome mistakes, and boosting their performance on complex, structured tasks like code generation and mathematical reasoning.



Table \ref{tab:tab3} presents a comparative analysis of accuracy on the HumanEval dataset among three different methods: APE, CoT, and our work (CD). The results demonstrate that CD consistently outperforms both APE and CoT across multiple models. For instance, GPT-3.5 shows an increase in accuracy from 0.93 with APE, 0.45 with CoT, but it observes the greatest lift to 0.96 with CD. Similarly, Claude-2 achieves near-perfect accuracy with CD at 0.99, compared to 0.96 with APE and 0.82 with CoT.

The results also highlight significant improvements for Mixtral-8x7B* and Mistral-7B, where CD boosts their accuracies to 0.95 and 0.96, respectively, compared to lower accuracies with APE (0.73 and 0.71) and CoT (0.88 and 0.87). Notably, Phi-3-mini-3.8B's accuracy slightly decreases with CD compared to CoT due to its initial weaknesses during training, which resulted in a lower baseline accuracy of 38\% on the training set. As a result, the extensive concept distillation required to address these weaknesses introduced slight confusion in some edge cases. Despite this, Phi-3-mini-3.8B still maintains competitive performance.

Table \ref{tab:tab4} provides a comprehensive comparison of different models and methods, including APE, CoT, Iterative APE, APO, PE2, and CD, across the various models on the Multi-Arith dataset. The results demonstrate that CD consistently outperforms other methods across most models. Particularly, GPT-3.5 achieves the highest accuracy with CD at 0.95, compared to 0.63 with APE and 0.71 with CoT. Similarly, Claude-2 shows a substantial improvement with CD, reaching an accuracy of 0.91, while other methods like APE and CoT achieve lower accuracies of 0.43 and 0.48, respectively.

Mixtral-8x7B* also benefits significantly from CD, achieving an accuracy of 0.88, compared to 0.84 with APE and 0.85 with CoT. However, Mistral-7B's performance slightly decreases with CD, achieving an accuracy of 0.67, compared to 0.72 with CoT. Similar to Phi-3-mini-3.8B in the previous section, we observed that the introduced concepts led to confusion for the Mistral-7B model on certain edge cases. Overall, Table \ref{tab:tab4} highlights the effectiveness of the CD framework, demonstrating its superior performance in enhancing model accuracy compared to other prompt optimization methods.

We also evaluated the transferability of optimized prompts from GPT-3.5 Turbo to other models like Claude 2.1, Phi-3-mini-3.8B, Mixtral-8x7B*, and Mistral-7B. Results show significant performance gains, with smaller models like Phi-3-mini-3.8B improving by 34\% and Claude 2.1 achieving 100\% accuracy. Detailed results and further analysis are provided in the Appendix \ref{subsec: concepts_transfer}. These findings highlight the generalizability of the distilled concepts across models. 

Finally, in Appendix \ref{sec: humanEval_benchmark}, we provide a qualitative analysis of the prompt changes generated for the HumanEval benchmark. This analysis demonstrates how CD extracts generalizable concepts to improve reasoning and adaptability in weak models, achieving substantial performance gains while addressing the limitations of rigid few-shot demonstrations.

%% file: 7conclusion_v1.tex
In conclusion, our study demonstrates the robustness of Concept Distillation in significantly enhancing the performance of weaker language models across various tasks, as evidenced by substantial accuracy improvements on the HumanEval, Multi-Arith, and GSM8K datasets. By distilling and transferring essential concepts from stronger models, CD not only boosts the capabilities of smaller models but also ensures the transferability of these improvements across different models. Our extensive experiments show that CD consistently outperforms various state-of-the-art prompt optimization methods. This robust framework, therefore, addresses critical challenges in prompt engineering, offering a scalable and resource-efficient solution that advances the state-of-the-art in prompt optimization for language models.

%% file: appendix.tex
\section{Notations and Prompt Templates}
\label{sec: Notations}

A detailed explanation of the notations used in Algorithm~\ref{alg:one} is presented in Table~\ref{tab:cd_notations}. The prompt templates are organized by the three phases of the algorithm: \textit{Initialization}, \textit{Induction}, and \textit{Deduction/Verification}, and are presented next. 

\begin{table*}[h]
    \centering
    \begin{tabular}{|p{2.8cm}|p{2.8cm}|p{9cm}|}
        \hline
        \textbf{Notation} & \textbf{Meaning} & \textbf{Description} \\
        \hline
        $M_s$ & Strong Model & The larger or more capable model (e.g. GPT-4o) used for generating and reasoning over concepts based on presented facts. \\
        \hline
        $M_w$ & Weak Model & The smaller or less capable model (e.g. Mistral-7B) whose performance on a given task is being optimized through concept distillation. \\
        \hline
        $D$ & Training Dataset & The dataset containing pairs of inputs ($x_i$) and expected outputs ($y_i$) used for assessing and optimizing the weak model performance on a given task. \\
        \hline
        $p_0$ & Initial Prompt & The base prompt template used as a starting point for the Weak Model before optimization. \\
        \hline
        $p$ & Updated Prompt & The prompt updated with distilled concepts during the optimization process.\\
        \hline 
        $x_i$ & Input Example & A single example from the training dataset used as input for the weak ($M_w$) and strong model ($M_s$).\\
        \hline 
        $y_i$ & Expected Output & The correct output corresponding to an input example, $x_i$. \\ 
        \hline
        $y_w$ & Weak Model's output & The output generated by the weak model for a given output example using the current prompt~$p$ in a given iteration of the CD process. \\
        \hline 
        $C$ & Set of Distilled Concepts & A collection of rules or concepts derived from the Strong Model that aim to address the Weak Model's deficiencies. \\
        \hline
        $R$ & Induced Concepts & Key concepts, rules, or examples generated by the Strong Model ($M_s$) during the induction phase to improve the Weak Model's performance. \\
        \hline
        $\texttt{ValidateConcept}$ & Concept Validation Function & A process that verifies the relevance and generalizability of the induced concepts $R$ based on validation set performance \\
        \hline
        $\texttt{InduceConcept}$ & Concept Induction Function & The function that leverages the strong model ($M_s$) to generate high-level, generalizable concepts from the failure reasons identified during the weak model's ($M_w$) evaluation \\
        \hline
        $\texttt{UpdatePrompt}$ & Prompt Update Function & A function that incorporates distilled concepts $C$ into the weak model's initial prompt ($p_0$) to create the updated prompt ($p$) which is then used for further evaluation \\
        \hline
    \end{tabular}
    \caption{CD Algorithm notations with their meanings and descriptions}
    \label{tab:cd_notations}
\end{table*}

\subsection{Initialization Prompt}
The initialization prompt ($p_0$) depends on the specific task. It can either be a baseline starting prompt or an existing production prompt for the weak model ($M_w$). The baseline prompt could be manually crafted or automatically generated to evaluate the weak model. An example of an initial prompt for a code generation task on HumanEval benchmark is shown in \ref{sec: humanEval_benchmark}.

\subsection{Induction Prompts}
The induction phase consists of two steps: \emph{(i)} generating the reasons for failures and \emph{(ii)} generating concepts. In both steps, the strong model ($M_s$) is used to identify the issues in the weak model's responses and then induce the concepts for improvements. These prompts take inputs such as the original task ($x_i$), the initial instruction prompt ($p_0$), the generated response ($y_w$) by the weak model, and the ground truth ($y_i$) to guide the process. The generated list of reasons for the weak model's failure from step 1 is also added to step 2's prompt to aid in the generation of key concepts.

\subsection{Deduction/Verification Prompts}
The deduction/verification phase refines the induced concepts ($R$) in order to minimize overfitting. This phase uses the strong model ($M_s)$ to analyze, and validate the induced concepts for the task before they are introduced into the weak model's ($M_w$) prompt $p$.

After refining and validating the induced concepts, an optional verification step is conducted. In this step, similar examples (\textit{task}) to the negative sample are selected either from the validation set or from synthetically generated examples using the strong model ($M_s$). The refined concepts are then introduced into the weak model's ($M_w$) prompt and tested against these similar examples. This step assesses whether the weak model can not only address the original mistake but also generalize to similar cases by achieving a predefined performance threshold. Only if this threshold is met are the refined concepts accepted as part of the final set of distilled concepts ($C$). The recommended threshold for this method is 80\%, ensuring that the weak model achieves consistent performance improvements across both the original mistake (negative sample) and similar examples.

\tcbset{
    mypromptstyle/.style={
        colback=blue!5,        
        colframe=cyan!55!black, 
        fonttitle=\bfseries,   
        coltitle=white,        
        colbacktitle=cyan!50!black,
        boxrule=1pt,           
        arc=4mm,               
        left=1mm,              
        right=1mm,             
        top=2mm,               
        bottom=3mm,            
        halign=flush left,
    }
}

\begin{tcolorbox}[title=Prompt for Induction Phase: Step 1 - Generate Reasons $\to  M_s$, mypromptstyle]
\textbf{\textless system\textgreater}  \\
    You are a skilled evaluator that can analyze instruction prompts and generated responses to identify issues.  
    For context, you will be given a task, an instruction prompt used to complete that task, a response to the task,  
    and the ground truth expected response. Your task is to identify reasons why the response failed to meet the ground truth.
    
    \vspace{0.5em}
    
    \textbf{\textless user\textgreater}  \\
    The original task is: \texttt{\{original\_task\}} \\  
    The instruction prompt used was: \texttt{\{instruction\_prompt\}} \\  
    The response generated based on the prompt is: \texttt{\{generated\_response\}} \\  
    An example of a correct ground truth is: \texttt{\{ground\_truth\}} \\  
    The evaluation result was: \texttt{\{evaluation\_result\}} \\ 
    
    \vspace{0.5em}
    
    Based on the evaluation result and the provided example ground truth,  
    can you identify a list of \texttt{\{n\}} reasons why the generated response failed?
\end{tcolorbox}

\begin{tcolorbox}[title=Prompt for Induction Phase: Step 2 - Generate Concepts $\to M_s$, mypromptstyle, top=2mm, bottom=1mm, breakable]
    \textbf{\textless system\textgreater}  \\
        You are a helpful assistant that can analyze instruction prompts and identify high-level, generalizable concepts 
        that can be added to the prompt to ensure the task is completed successfully.  
        A concept is a general instruction derived or inferred from specific instances or occurrences.  
        Concepts should be general enough to be applicable to a wide range of tasks.
    
    \vspace{0.5em}
    
    \textbf{\textless user\textgreater}  \\
        - The original instruction prompt was: \texttt{\{original\_prompt\}} \\  
        - The response was: \texttt{\{generated\_response\}} \\  
        - The ground truth expected response was: \texttt{\{ground\_truth\}} \\  
        - Reasons for the failure include: \texttt{\{failure\_reasons\_step\_1\}} \\  
    
    \vspace{0.5em}
    
    Can you identify a list of \texttt{\{n\}} concepts that can be added to the prompt to ensure the task as well as related ones passes?
    
    \vspace{0.5em}
    
\end{tcolorbox}



\begin{tcolorbox}[title=Deduction Phase: Refine and Filter Concepts $\to M_s$, mypromptstyle, colback=lightgray!12, colbacktitle=lightgray!50!black, colframe=lightgray!50!black]
    \textbf{\textless system\textgreater}  \\
        You are an intelligent assistant that processes a list of high-level, generalizable concepts for a given task. Your task is twofold: \\
        1. Analyze the list of concepts and remove semantically similar duplicates, ensuring that each remaining concept is unique and distinct. \\
        
        2. Verify that each concept is general enough to be valid for improving the given task. A valid concept should: \\
        \begin{itemize}
            \item Be generalizable to similar examples within the task.
            \item Directly address weaknesses or improve performance for the task.
        \end{itemize}

        A concept is defined as a general instruction derived or inferred from specific instances or occurrences of a task. Your goal is to preserve the clearest, most concise, and generalizable version of each valid concept.
    
    \vspace{0.5em}
    
    \textbf{\textless user\textgreater}  \\
        Here is the list of concepts generated for the task: \texttt{\{concepts\}} \\  
        The original task is: \texttt{\{original\_task\}} \\
        
    \vspace{0.5em}
    
    Please return a list of unique, valid concepts. Your output should only include the refined concepts without any additional explanations or preambles.
\end{tcolorbox}

During the verification process, if a newly introduced concept does not contribute to a measurable performance improvement, it is more likely to be discarded. This ensures that only useful concepts are retained, effectively filtering out detrimental refinements. Redundant concepts, on the other hand, are handled explicitly through instructions provided in the deduction phase prompt, which ensure that semantically similar concepts are merged or eliminated while preserving generalizability. By combining empirical validation with structured filtering mechanisms, the framework optimally refines distilled concepts without compromising useful knowledge.

\begin{tcolorbox}[title=Updated Prompt Template for Verification $\to M_w$, mypromptstyle, colback=lightgray!12, colbacktitle=lightgray!50!black, colframe=lightgray!50!black]
    \textbf{\textless system\textgreater}  \\
    You are a helpful assistant that performs \texttt{\{task\}}.  
    Follow the given instructions to complete the task successfully.
    
    \vspace{0.5em}
    
    \textbf{\textless user\textgreater}  \\
    Key concepts to follow: \texttt{\{key\_concepts\}} \\
    
    Instructions: \texttt{\{initial\_prompt\}}
\end{tcolorbox}


\section{Additional Results}
\label{sec:additional_results}

In this section, we provide further quantitative and qualitative results complementing our experiments.

\subsection{GSM8K Dataset} \label{sec: additional_GSM8K}
Table \ref{tab:tab5} presents the accuracy comparison on the GSM8K dataset between CD and APE. The results demonstrate that CD consistently outperforms APE across multiple models. For instance, the GPT-3.5 model shows a significant improvement in accuracy from 0.67 with APE to 0.76 with CD. Similarly, GPT-4's accuracy increases from 0.84 with APE to 0.90 with CD, highlighting the effectiveness of CD in enhancing model performance on mathematical reasoning tasks.

Despite these improvements, the Claude 2.1 model experienced a slight decrease in performance, dropping from 0.86 with APE to 0.84 with CD. This suggests that while CD is generally effective, it may introduce prompt overload that can sometimes negatively impact certain models, particularly in scenarios involving highly comprehensive datasets like GSM8K.
Future work will explore methods to encourage the consolidation of distilled concepts or the development of a hierarchical structure of concepts to enhance their effectiveness. 

\begin{table}[ht]
    \centering
    \begin{tabularx}{0.5\textwidth}{c|X|X}
    \hline
    \textbf{Model} & \textbf{APE} & \textbf{CD} \\
    \hline
        GPT-3.5 & 0.67 & \textbf{0.76} \\
    \hline
        Claude 2.1 & \textbf{0.86} & 0.84   \\
    \hline
        GPT-4 & 0.84 & \textbf{0.90} \\
    \hline
    \end{tabularx}
    \caption{Accuracy comparison on the GSM8K dataset between CD and APE}
    \label{tab:tab5}
\end{table}

\subsection{Transferability of Distilled Concepts}
\label{subsec: concepts_transfer}
We tested how well the optimized prompts, originally designed for GPT-3.5 Turbo, work on other models like Claude 2.1, Phi-3-mini-3.8B, Mixtral-8x7B*, Mistral-7B, and GPT-4. This helped us see if the distilled concepts are effective across different language models.

Table \ref{tab:transf_1} provides compelling evidence for our hypothesis that distilled concepts from CD are transferable and generalizable across different models. In this experiment, GPT-3.5 Turbo served as the base model for distilling concepts using a strong model (GPT-4o), and the optimized prompts were then transferred to other models for evaluation. We observe significant performance improvements across all models. Notably, Claude 2.1 achieved a perfect score of 100\%, demonstrating an 11\% improvement. The smallest model, Phi-3-mini-3.8B, exhibited the most remarkable improvement, with a performance boost of 34\%, increasing its accuracy from 0.45 to 0.79. This result further validates the observation that smaller models gain substantial benefits from the CD process. Overall, the results show an average performance increase, confirming that the distilled concepts are not only effective for the base model but also enhance the performance of other models significantly.

\begin{table}[h]
    \centering
    \begin{tabularx}{0.5\textwidth}{c|X|X}
    \hline
    \textbf{Model} & \textbf{Base prompt} & \textbf{CD} \\
    \hline
        GPT-3.5 & 0.85 & \textbf{0.96$_{(+11\%)}$} \\
    \hline
        Claude 2.1 & 0.89 & \textbf{1.00$_{(+11\%)}$}  \\
    \hline
        Phi-3-mini-3.8B & 0.45 & \textbf{0.79$_{(+34\%)}$} \\
    \hline
        Mixtral-8x7B* & 0.83 & \textbf{0.87$_{(+5\%)}$} \\
    \hline
        Mistral-7B & 0.89 & \textbf{0.96$_{(+7\%)}$} \\
    \hline
        GPT-4 & 0.90 & \textbf{0.94$_{(+4\%)}$} \\
    \hline
    \end{tabularx}
    \caption{Accuracy results on the HumanEval dataset: The results demonstrate the effectiveness of transferring an optimized prompt (with distilled concepts) based on the GPT-3.5-Turbo model to other models}
    \label{tab:transf_1}
\end{table}

Table \ref{tab:transf_2} provides a comparative analysis of the accuracy improvements achieved through distilled concepts transfer from GPT-3.5 Turbo prompt optimized using CD to both smaller and larger models, compared to APE on the HumanEval dataset. The CD method significantly outperforms APE, with notable improvements in models such as Mistral-7B, which saw a substantial increase of 25\% (from 0.71 to 0.96). Mixtral-8x7B* also benefited greatly, with a 14\% boost in accuracy (from 0.73 to 0.87). These results show the superior performance of the CD approach in enhancing model performance by distilling and transferring essential concepts from stronger to weaker models.

\begin{table*}[h]
\normalsize
    \centering
    \begin{tabular}{|c|l|l|l|l|l|l|}
        \hline
        \textbf{Method} & \textbf{GPT-3.5} & \textbf{Claude 2.1} & \textbf{Phi-3-mini-3.8B} & \textbf{Mixtral-8x7B*} & \textbf{Mistral-7B} & \textbf{GPT-4} \\ \hline
        APE & 0.93 & 0.96 & \textbf{0.83$_{(+4\%)}$} & 0.73 & 0.71 & 0.91 \\ \hline
        \textbf{CD} & \textbf{0.96$_{(+3\%)}$} & \textbf{1.00$_{(+4\%)}$} & 0.79 & \textbf{0.87$_{(+14\%)}$} & \textbf{0.96$_{(+25\%)}$} & \textbf{0.94$_{(+3\%)}$} \\ \hline
    \end{tabular}
    \caption{Accuracy comparison on the HumanEval dataset between CD (evaluated by transferring the optimized prompt with distilled concepts from the GPT-3.5-Turbo model to other models) and APE.}
    \label{tab:transf_2}
\normalsize
\end{table*}


\subsection{Qualitative analysis for the HumanEval Benchmark} \label{sec: humanEval_benchmark}
Below, we present the simple prompt initially used for completing the HumanEval task, followed by the optimized prompt enriched with distilled concepts for GPT-3.5-Turbo from the HumanEval benchmark. The optimized prompt includes specific examples of distilled concepts that highlight CD's ability to generalize and improve model performance.

\begin{figure}[h]
    \centering
    \includegraphics[width=1.0\linewidth]{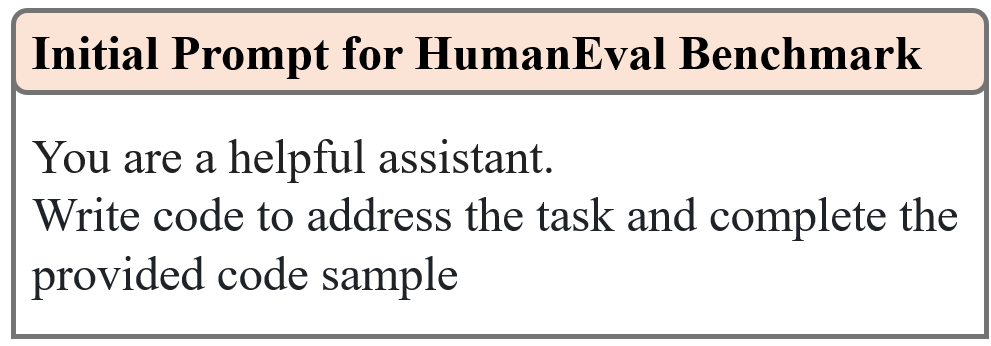}
\end{figure}

\begin{figure}[h]
    \centering
    \includegraphics[width=1.0\linewidth]{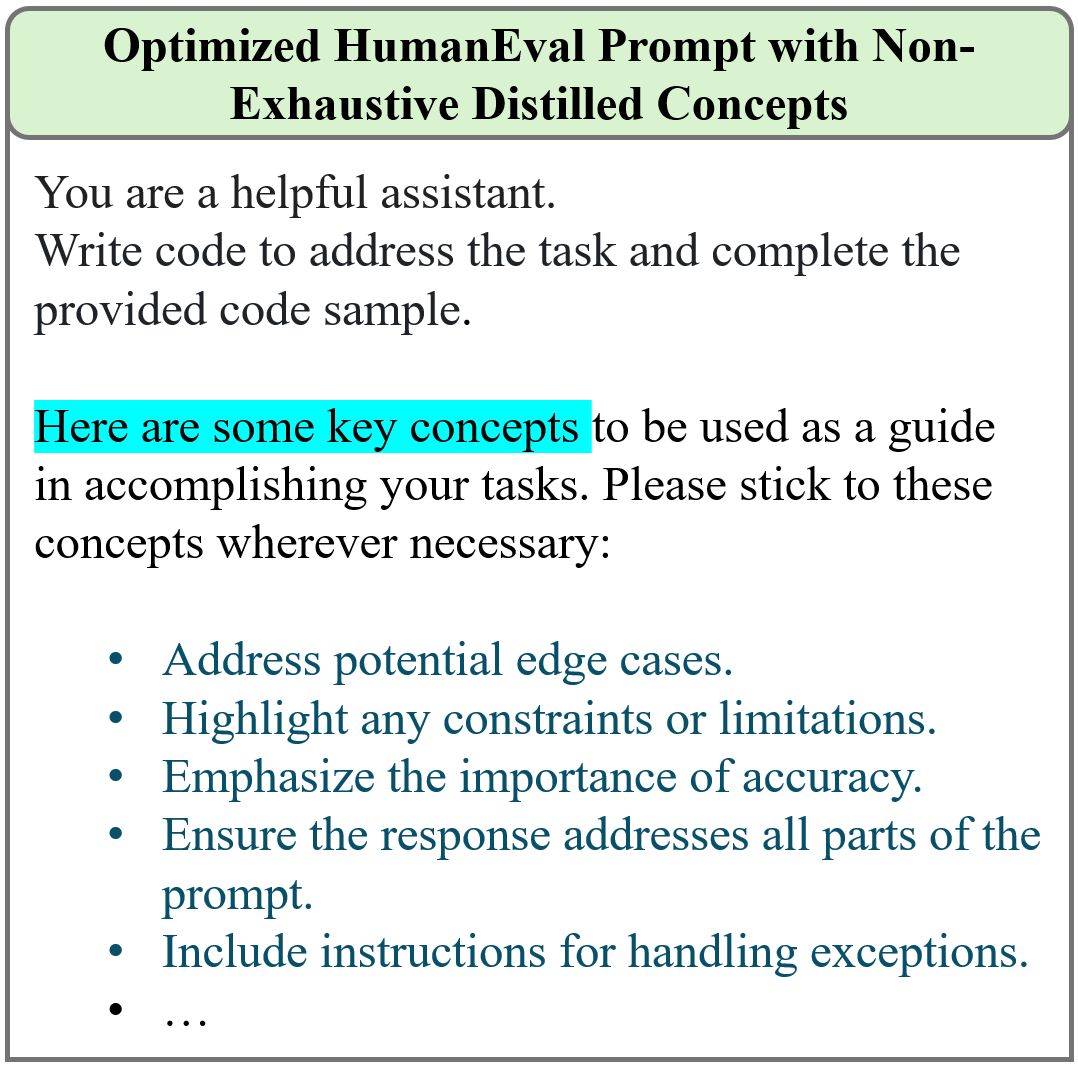}
\end{figure}

As shown in the optimized prompt for the case of HumanEval benchmark above, these distilled concepts are insightful yet concise concepts that address several non-trivial dimensions of the problem at hand. The distilled concepts ensure that explicit constraints, such as ensuring type compatibility in arithmetic operations, are enforced to minimize errors. Furthermore, our additional experiments (refer to Appendix~\ref{subsec: concepts_transfer}) demonstrate the transferability of distilled concepts from GPT-3.5 Turbo to other models such as Claude 2.1, Phi-3-mini-3.8B, Mixtral-8x7B, and Mistral-7B. Results show that Phi-3-mini-3.8B improved by 34\%, while Claude 2.1 achieved 100\% accuracy on key benchmarks. These findings indicate that the distilled concepts enable weaker models to perform well on complex reasoning tasks, thus validating that CD introduces meaningful reasoning improvements beyond simple formatting error corrections.

\section{Natural Language to Cypher Translation: Case Study}
\label{sec: hypothetical_example}

In this section, we present an industry case study covering a task aiming to translate natural language queries to a graph database query language (Cypher).

\subsection{Walk-through of the method} 
To illustrate our proposed method, we employ a hypothetical example, guiding you through the three phases of the concept distillation process shown in Fig. \ref{fig:intro_CD}. 

\begin{figure}[h]
    \centering
    \includegraphics[width=1.1\linewidth]{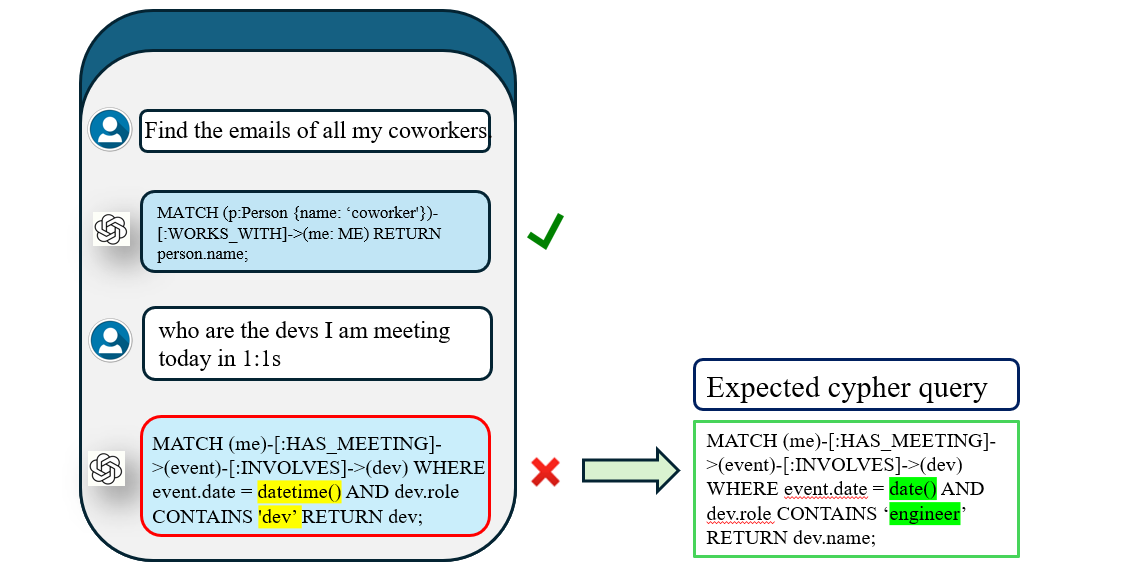}
    \caption{A hypothetical natural language to cypher query translator}
    \label{fig:hypothetical_example}
\end{figure}

The example involves a chatbot designed to translate natural language into Cypher query commands. \textit{Cypher} is a declarative graph query language used for querying and managing data in graph databases, such as Neo4j. It enables users to efficiently and intuitively query, update, and manage graph data by expressing patterns in the graph structure through a readable syntax. This chatbot utilizes an LLM, specifically GPT-3.5, to interpret a user's natural language query and generate a corresponding Cypher query based on a predefined graph schema. This example will demonstrate how our technique optimizes the prompt of the assumed weak model in question (GPT-3.5). Figure \ref{fig:hypothetical_example} depicts the hypothetical natural language to Cypher query translator utilized for the purpose of explaining the method.

\subsubsection{Initialization}
In this initial phase, we set up essential components for our technique. This includes defining the task (natural language to Cypher query translation in this case), preparing a 'golden dataset' which contains pairs of natural language queries and their corresponding Cypher queries (serving as the task's ground truth), and creating a prompt template with basic information and instructions for the task. This template might include a few examples, and specify the input-output format. The golden dataset represents the training dataset for the method. Depending on the size of the training dataset set, we cluster it into various entities and then use stratified sampling technique to split the dataset into train and validation sets  The initial task-specific prompt used for this phase could be generated by an off-the-shelf algorithm, manually crafted, or an already existing prompt being used by a different LM.

\begin{figure*}[h]
    \centering
    \includegraphics[width=0.9\linewidth]{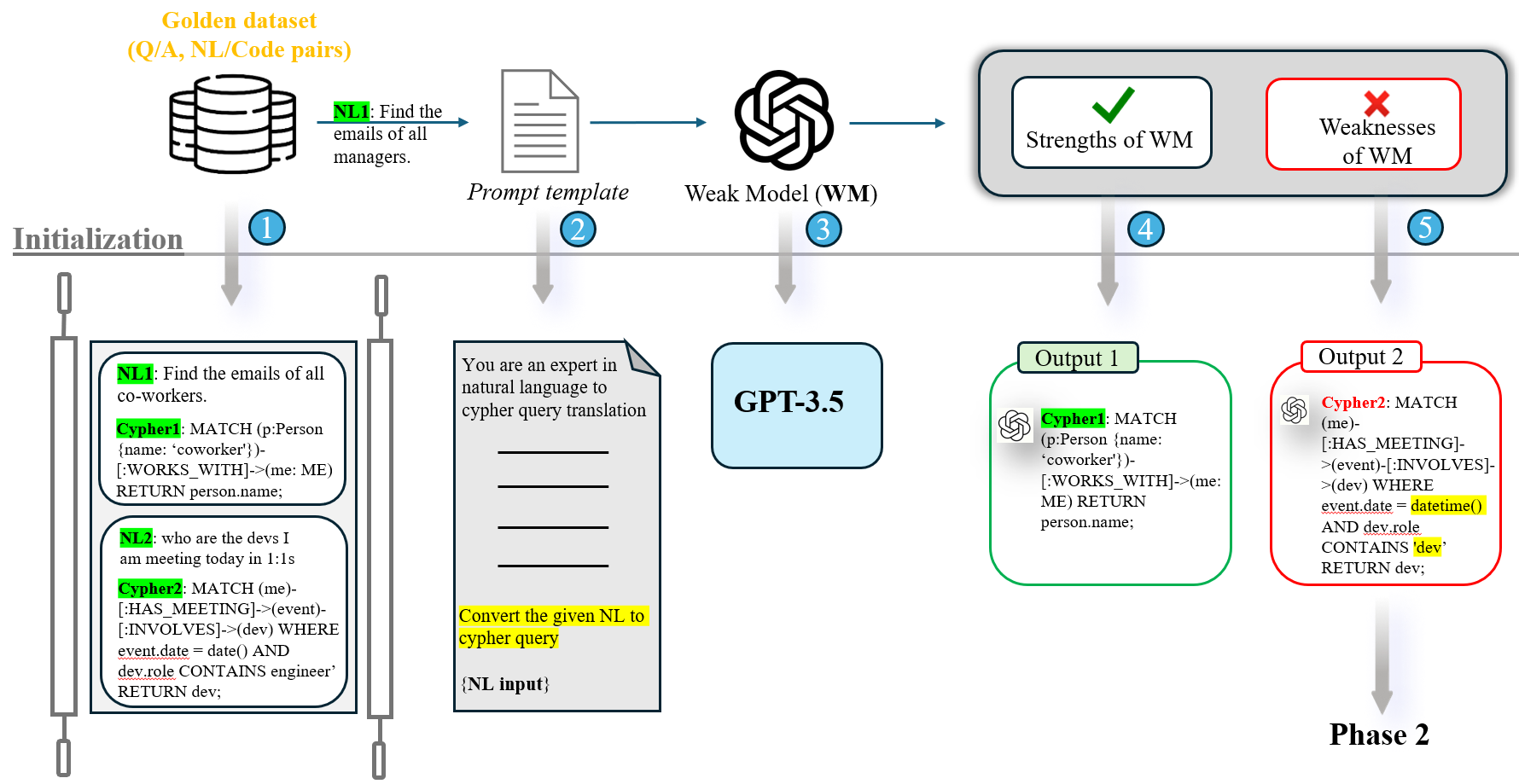}
    \caption{Initialization phase of concept distillation}
    \label{fig:cd_initialization}
\end{figure*}

We then evaluate the weak model, in this case, GPT-3.5, using this golden dataset of NL-Cypher pairs, as illustrated in Fig. \ref{fig:cd_initialization}. We start by selecting a pair of natural language and Cypher queries from the dataset and feeding them to the weak model using the prompt template. We then observe the output of the weak model and compare it to the ground truth. If the output is correct, we move on to the next pair. If the output is wrong, we record the error and proceed to the next step. In our hypothetical example, the first data point is deemed a strength of GPT-3.5 as it correctly generates the expected Cypher query. However, the second data point reveals a weakness, with the model failing to generate the correct Cypher query in response to the natural language query “who are the devs I am meeting in 1:1s.”

\subsubsection{Induction}
In this phase, we use the strong model to induce key concepts and rules from the given task and dataset, by prompting it to reason through the facts. Here, we start constructing the prompt for the strong model by going through the following steps:
\begin{itemize}
    \item First, we define the persona of the strong model, for example, “you are an expert in generating and reasoning over natural language to Cypher queries translation…”
    \item Next, we present to the strong model the accurate NL-Cypher pair - specifically, the one that the weak model failed to predict correctly. Along with this, we include in the strong model's prompt the incorrect Cypher query generated by the weak model, as well as the original prompt template that was used for the weak model.
    \item Following this, we request the strong model to analyze and identify the reasons behind the weak model's incorrect response. This analysis is based on all the information and facts that have been included in the prompt.
    \item The strong model then reasons through the facts presented and tries to provide a sense of meaning into why the weak model is struggling with the input query, which we are considering in this case as “who are the devs I am meeting in 1:1s.” Here, we ask the strong model to explain why the response of the weak model is wrong, and what are the missing or incorrect concepts or rules that the weak model should have used.
    \item We then follow up with another turn of discussion, in this case, we prompt the strong model to induce some concepts (concepts here could be rules, examples, etc. depending on the application) to guide the weak model in explicit reasoning, in such a way that it is able to answer all similar questions correctly.
    \item The strong model finally induces these concepts based on the presented facts and its reasoning over the cause of the weak model’s inability to generate the correct response.
\end{itemize}

\begin{figure*}[h]
    \centering
    \includegraphics[width=0.9\linewidth]{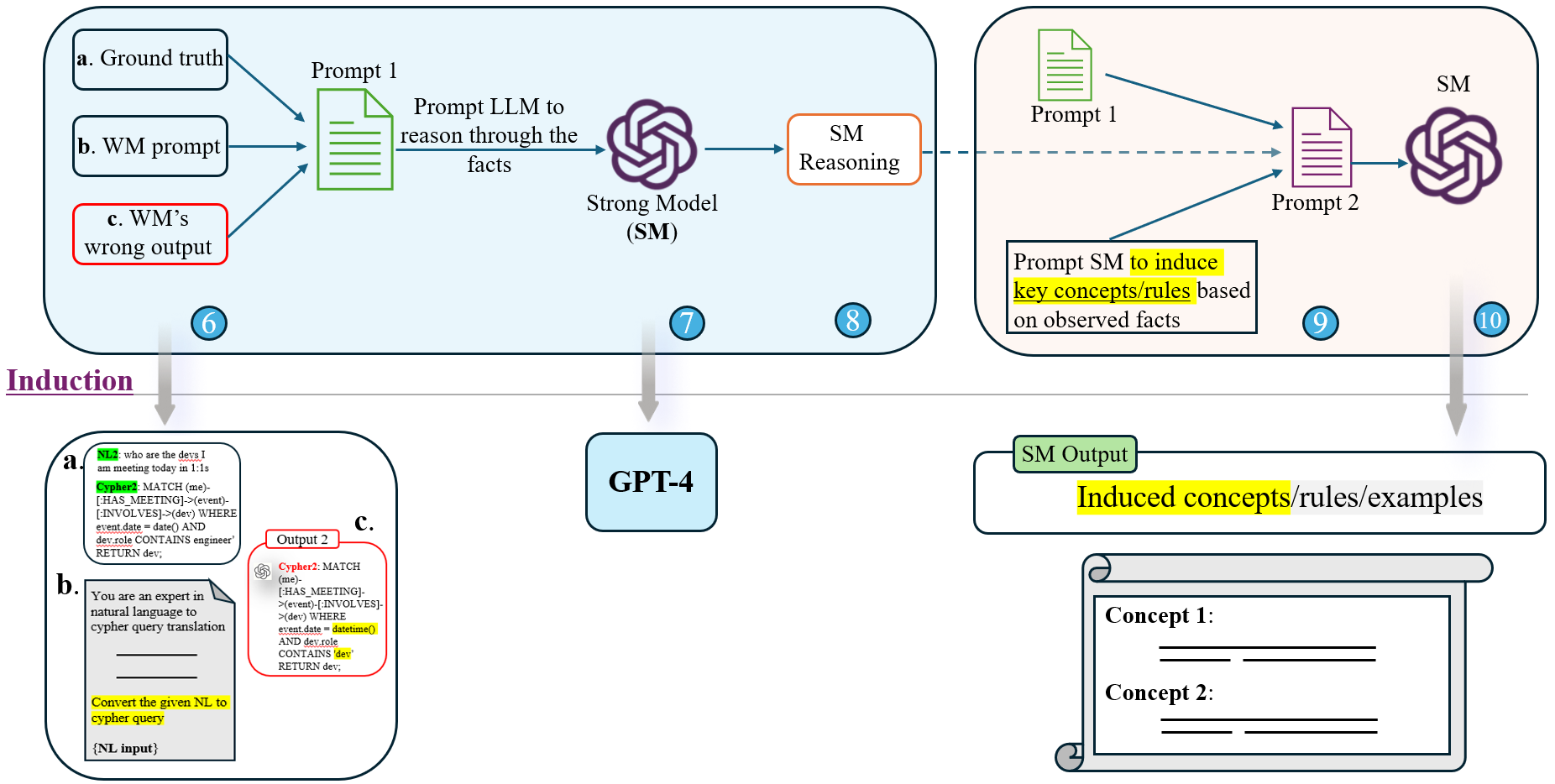}
    \caption{Induction phase of concept distillation}
    \label{fig:cd_induction}
\end{figure*}

Figure \ref{fig:cd_induction} illustrates the induction phase of the concept distillation method. As noted earlier in the preceding sections, not all induced concepts are general enough to be considered as distilled concepts and so we got through the final step of this approach, which is deduction from verification, to verify these concepts to either accept or reject them. 

\begin{figure*}[t]
    \centering
    \includegraphics[width=0.9\linewidth]{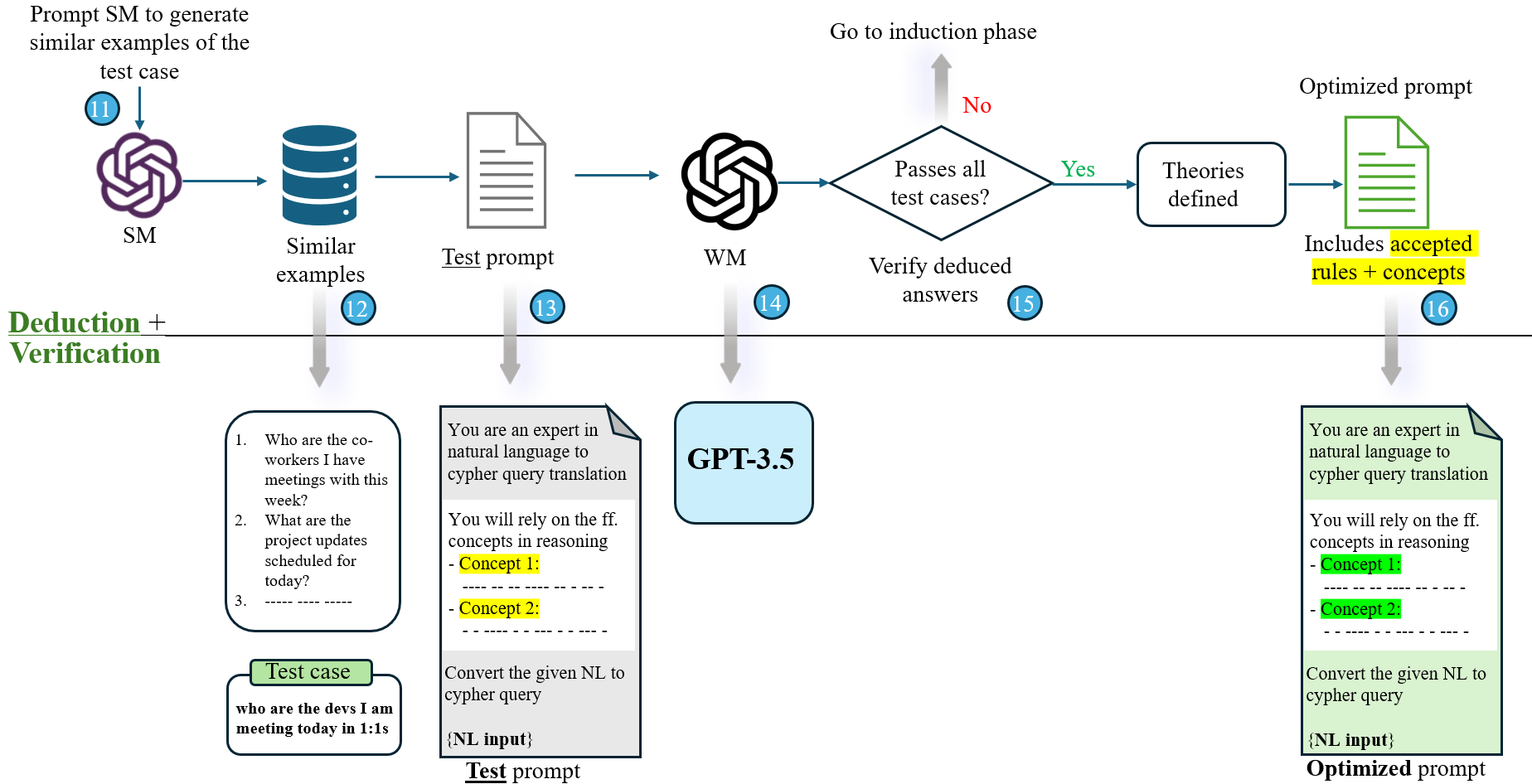}
    \caption{Deduction from Verification phase of concept distillation}
    \label{fig:cd_deduction}
\end{figure*}

\subsubsection{Deduction from Verification}
The final phase, \textit{Deduction from Verification}, employs deductive reasoning to validate the concepts induced in the previous phase. This involves using the strong model to generate test cases that are similar to the incorrectly predicted input in questions (as “who are the devs I am meeting in 1:1s.”). The generated test cases mimic the initial failure but with varied contexts or phrasings. \textit{Alternatively}, a sample from the same entity in the validation dataset that closely resembles this test case could be used for this process. Similar examples generated in this scenario could be “Who are the co-workers I have meetings with this week?”, and “What are the project updates scheduled for today?” We then observe the output of the strong model, and select a subset of the generated examples that are valid and relevant for the task.

Following this, we incorporate the concepts derived from the induction phase into the weak model's (GPT-3.5) original prompt template, creating what we'll refer to as the 'test prompt.' Using this test prompt, we then re-evaluate the weak model on both the original incorrectly predicted input and all the newly generated similar examples. The aim is to verify whether the model's responses, now informed by the revised test prompt, correctly align with the expected answers. If the weak model is now able to deduce correct responses for all test examples with a level of certainty or probability that meets or exceeds a specific predefined threshold, the we have our theories defined and as a result, we can go ahead and accept the induced concepts as distilled concepts; otherwise, the induced concepts are rejected and we go back to the induction phase, and generate more concepts and rules from the strong model, until the weak model passes all the test cases. Figure \ref{fig:cd_deduction}

We repeat this process for different pairs of queries that the weak model struggles with until we have a sufficient number of distilled concepts for the task, that can significantly boost the performance of the weak model for the task-specific domain. In practice, an intriguing observation we have made is that distilling concepts for one specific negative sample in the golden dataset often corrected not only that particular sample but also other negative samples where the weaker model had previously failed.

This iterative process of distilling concepts from a strong model to a weak model forms the cornerstone of our methodology. It enables a precise, targeted enhancement of the weak model's capabilities, addressing specific deficiencies with tailored improvements. Through this approach, we not only rectify isolated errors but also fortify the model's overall performance for the given task.

\subsection{Quantitative Analysis}

In this study, we also employed the concept distillation approach on a dataset designed for Natural Language to Cypher (NL2Cypher) query translation, aiming to leverage the generative capabilities of LLMs for producing syntactically correct Cypher codes from natural language queries. The dataset encompassed various subsets, including queries pertaining to calendars (e.g., "when is my next meeting with person"), files, and people, structured according to a specific schema.

\begin{figure}[h]
    \centering
    \includegraphics[width=1.0\linewidth]{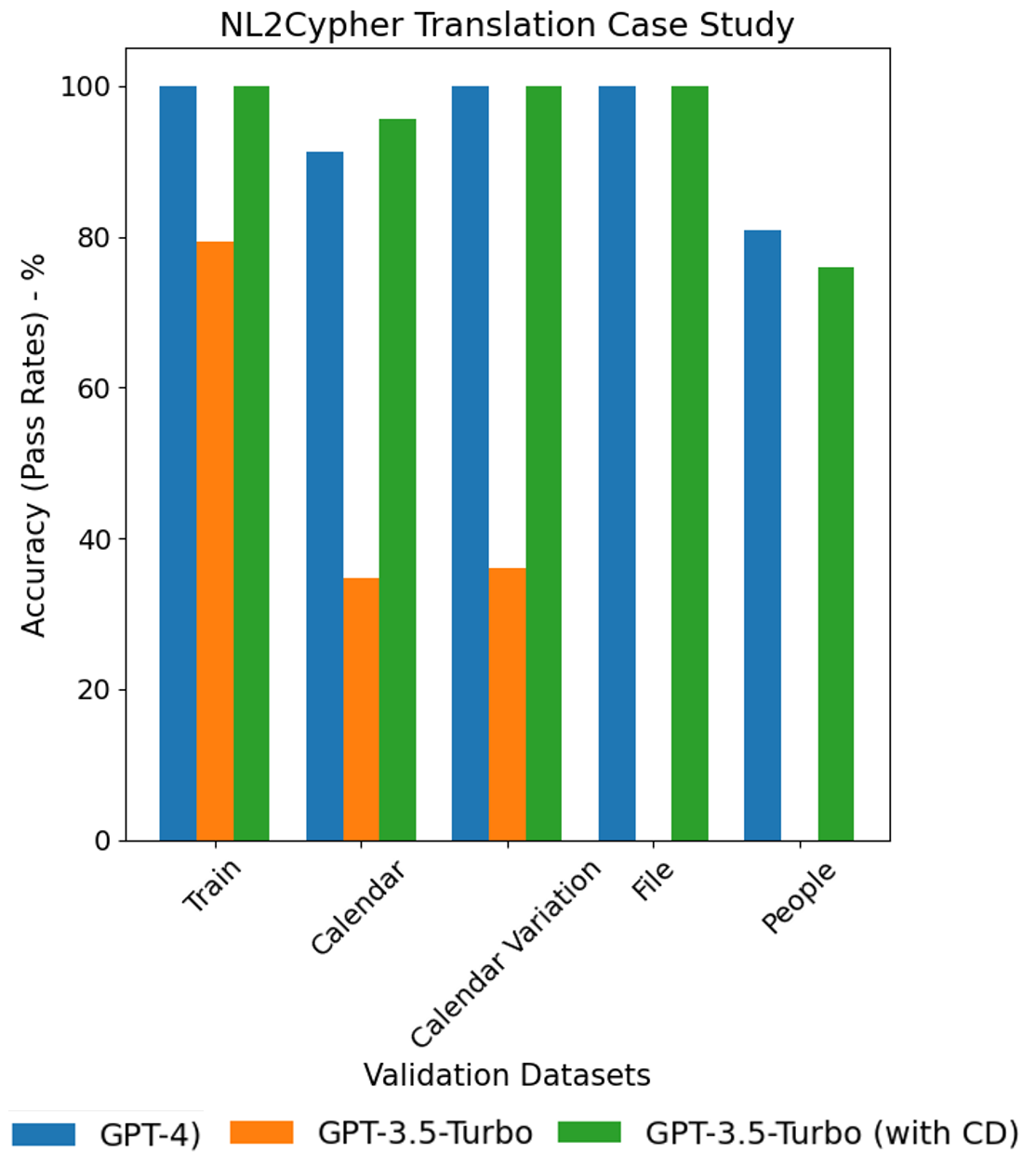}
    \caption{Accuracy (pass rate) comparison between GPT-3.5 Turbo model (with and without CD) and GPT-4}
    \label{fig: nl2cypher_result}
\end{figure}

Our observations highlighted that the GPT-4 model demonstrated superior performance across all dataset subsets during validation, with its lowest accuracy—approximately 80\%—occurring in NL2Cypher query translations concerning people. Conversely, the GPT-3.5 Turbo model, utilizing identical prompts to GPT-4, exhibited markedly lower performance across these subsets. Notably, it failed entirely to translate queries related to files and people within an organization, resulting in zero accuracy for these categories. Figure \ref{fig: nl2cypher_result} shows the accuracy comparison between GPT-3.5-Turbo (with baseline prompt), GPT-3.5 Turbo model (with optimized prompt) and GPT-4 model (with baseline prompt).

Subsequent to the application of concept distillation from GPT-4 into the prompt optimization process for GPT-3.5 Turbo—the performance of the latter model saw substantial improvements across all validation datasets. In particular, for queries related to the calendar category, the GPT-3.5 Turbo model not only improved but also exceeded GPT-4's performance, achieving an accuracy rate of 95.65\%. Moreover, in scenarios involving people-related queries, where the GPT-3.5 Turbo model initially failed to translate correctly any query, the incorporation of distilled concepts significantly enhanced its accuracy to approximately 76\%. For the GPT-3.5 Turbo model, the optimization of the prompt involved exclusively the incorporation of distilled concepts, resulting in what is termed the "optimized prompt." This approach demonstrates how the process of concept distillation can effectively guide a weaker model to regress towards the expected output during ICL.

\subsection{Qualitative Analysis}
In this section, we present a qualitative analysis of CD's behavior in comparison to conventional few-shot demonstrations for the NL2Cypher case study. By examining the limitations of few-shot demonstrations and comparing them to CD’s approach, we illustrate how CD enhances generalization and improves reasoning. 

In this work, we initially started with a baseline prompt which did contain few-shot demonstrations, an example of which is shown below, with about 125 tokens:

\begin{figure}[h]
    \centering
    \includegraphics[width=1.0\linewidth]{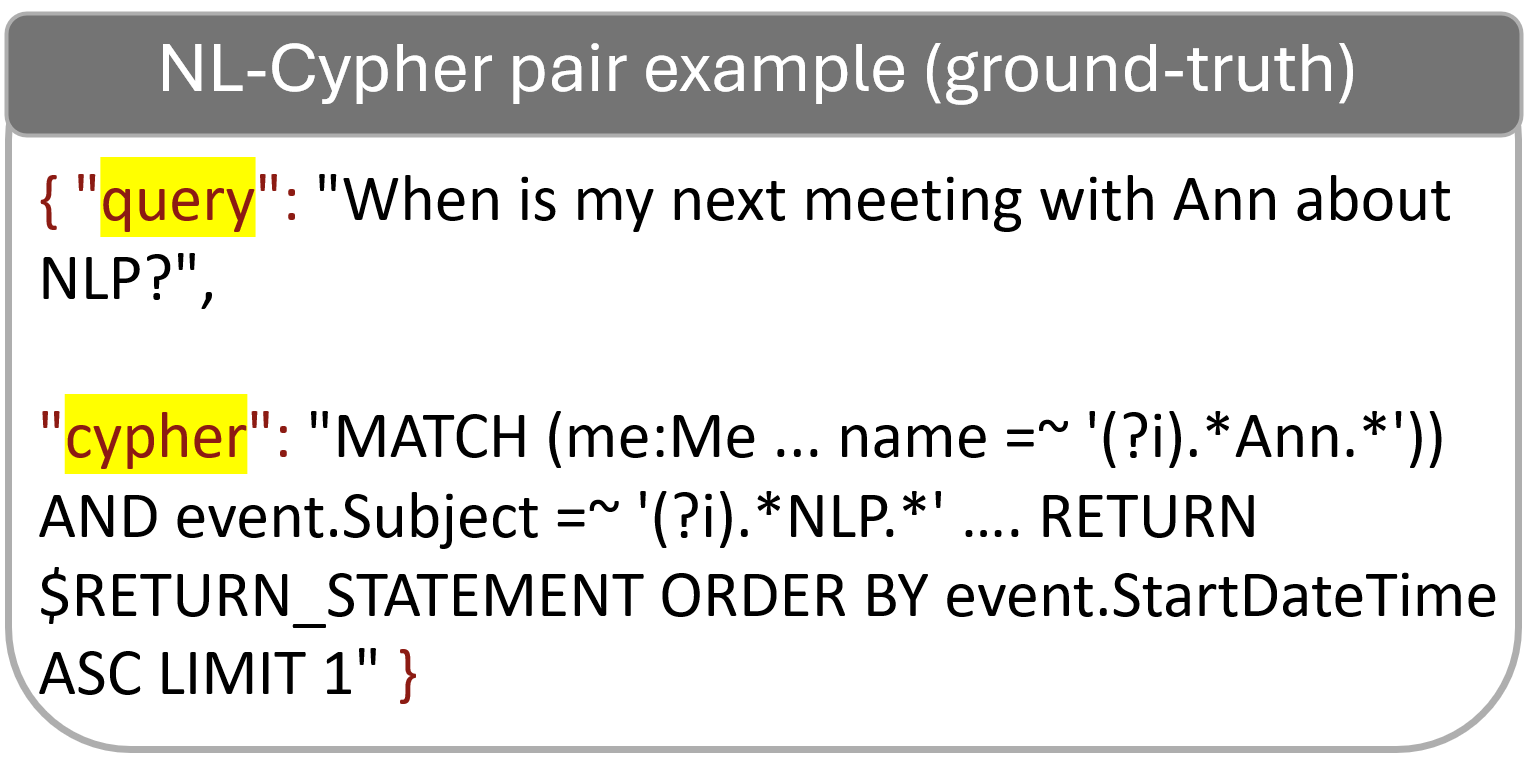}
\end{figure}

The specificity of these few-shot demonstrations in the prompt led to poor performance across several benchmarks due to its lack of generalization to different entities. The weaker model (in this case, GPT-3.5 Turbo model) tended to overfit to such specific scenarios, limiting its reasoning ability when handling other queries with different entity mentions.

In contrast, by applying CD, we distilled general, high-level concepts that helped the weaker model understand how to utilize demonstrations in a more flexible and general way. For example, one distilled concept for this case study took the forms of an improved example:
\begin{figure}[h]
    \centering
    \includegraphics[width=1.0\linewidth]{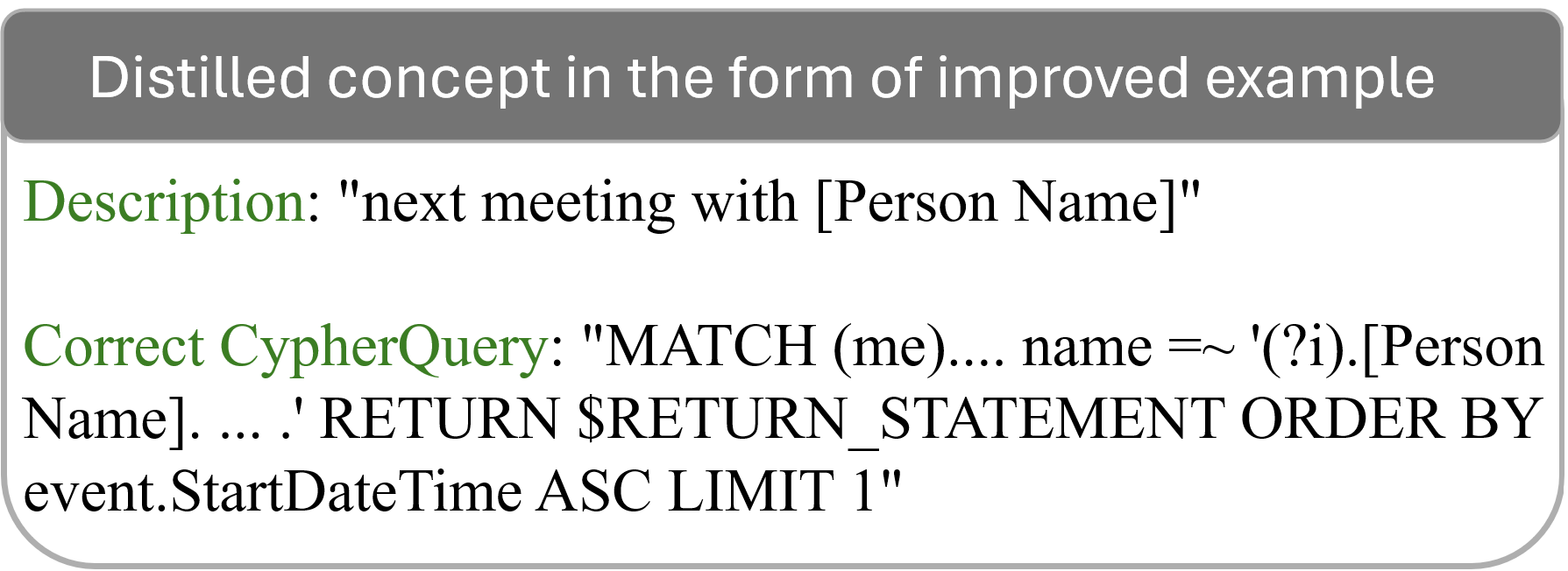}
\end{figure}

In this case, the distilled concept abstracts away the specifics of the demonstration by introducing a placeholder, \textit{[Person Name]} which can dynamically accommodate any person's name. The corresponding Cypher query also uses similar placeholder logic, enabling it to match to any name. This makes the distilled concept generalizable, enabling the weaker model to apply the same reasoning to a wide variety of queries involving different entity mentions without overfitting to specific examples or requiring additional examples for each case. Other concepts took the form of rules at different form of generality, as can be seen for the three examples below:

\begin{figure}[h]
    \centering
    \includegraphics[width=1.0\linewidth]{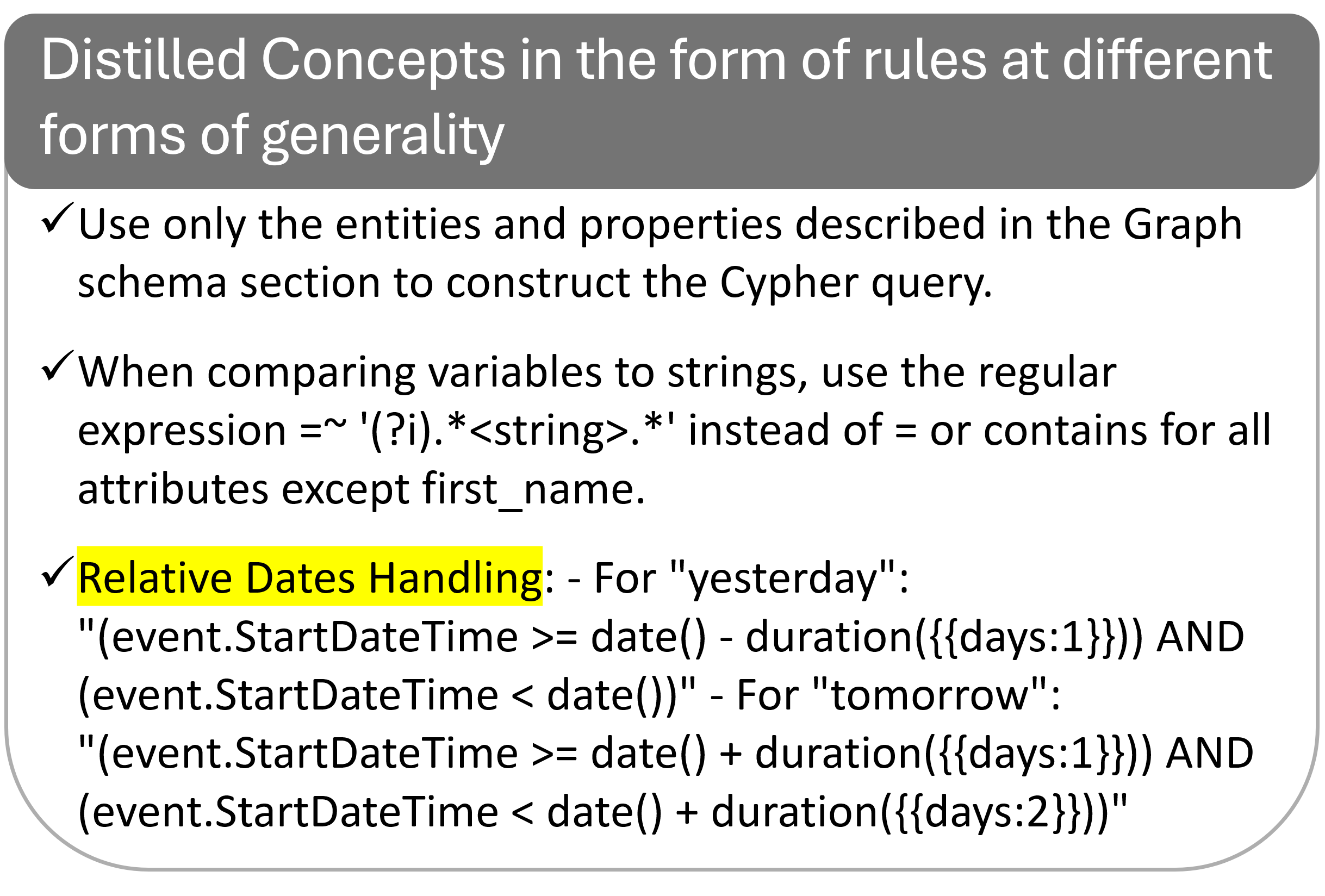}
\end{figure}

Also, the above four concepts had a smaller token footprint: 96, 19, 32, and 74 tokens respectively. By employing CD in this case study, we observed significant performance improvements across all benchmarks, including increase in pass rate from 0\% with few shot demonstrations to 100\% with distilled concepts as shown in Fig. \ref{fig: nl2cypher_result}. This shows how CD enhances the weaker model’s reasoning ability by providing it with general, reusable rules instead of rigid demonstrations.

This practical case demonstrates how CD offers a more efficient and scalable solution that complements adding specific demonstrations, both in terms of token cost and performance improvements.

%% file: acl_latex.bbl
\begin{thebibliography}{33}
\providecommand{\natexlab}[1]{#1}

\bibitem[{Bradford and Hamer(2022)}]{bradford2022science}
Alina Bradford and Ashley Hamer. 2022.
\newblock Science and the scientific method: Definitions and examples.
\newblock \emph{Published January}, 17:2022.

\bibitem[{Chen et~al.(2021)Chen, Tworek, Jun, Yuan, Pinto, Kaplan, Edwards, Burda, Joseph, Brockman et~al.}]{chen2021evaluating_humaneval}
Mark Chen, Jerry Tworek, Heewoo Jun, Qiming Yuan, Henrique Ponde de~Oliveira Pinto, Jared Kaplan, Harri Edwards, Yuri Burda, Nicholas Joseph, Greg Brockman, et~al. 2021.
\newblock Evaluating large language models trained on code.
\newblock \emph{arXiv preprint arXiv:2107.03374}.

\bibitem[{Cherukunnath and Singh(2022)}]{cherukunnath2022exploring}
Deepa Cherukunnath and Anita~Puri Singh. 2022.
\newblock Exploring cognitive processes of knowledge acquisition to upgrade academic practices.
\newblock \emph{Frontiers in Psychology}, 13:682628.

\bibitem[{Cobbe et~al.(2021)Cobbe, Kosaraju, Bavarian, Chen, Jun, Kaiser, Plappert, Tworek, Hilton, Nakano et~al.}]{cobbe2021training}
Karl Cobbe, Vineet Kosaraju, Mohammad Bavarian, Mark Chen, Heewoo Jun, Lukasz Kaiser, Matthias Plappert, Jerry Tworek, Jacob Hilton, Reiichiro Nakano, et~al. 2021.
\newblock Training verifiers to solve math word problems.
\newblock \emph{arXiv preprint arXiv:2110.14168}.

\bibitem[{Deng et~al.(2023)Deng, Zhang, Chen, and Gu}]{deng2023rephrase}
Yihe Deng, Weitong Zhang, Zixiang Chen, and Quanquan Gu. 2023.
\newblock Rephrase and respond: Let large language models ask better questions for themselves.
\newblock \emph{arXiv preprint arXiv:2311.04205}.

\bibitem[{Dong et~al.(2022)Dong, Li, Dai, Zheng, Wu, Chang, Sun, Xu, and Sui}]{dong2022survey}
Qingxiu Dong, Lei Li, Damai Dai, Ce~Zheng, Zhiyong Wu, Baobao Chang, Xu~Sun, Jingjing Xu, and Zhifang Sui. 2022.
\newblock A survey on in-context learning.
\newblock \emph{arXiv preprint arXiv:2301.00234}.

\bibitem[{Edwards and Camacho-Collados(2024)}]{edwards2024language}
Aleksandra Edwards and Jose Camacho-Collados. 2024.
\newblock Language models for text classification: Is in-context learning enough?
\newblock \emph{arXiv preprint arXiv:2403.17661}.

\bibitem[{Guo et~al.(2023)Guo, Wang, Guo, Li, Song, Tan, Liu, Bian, and Yang}]{guo2023connecting_evoprompt}
Qingyan Guo, Rui Wang, Junliang Guo, Bei Li, Kaitao Song, Xu~Tan, Guoqing Liu, Jiang Bian, and Yujiu Yang. 2023.
\newblock Connecting large language models with evolutionary algorithms yields powerful prompt optimizers.
\newblock \emph{ICLR 2024}.

\bibitem[{Hadi et~al.(2023)Hadi, Qureshi, Shah, Irfan, Zafar, Shaikh, Akhtar, Wu, Mirjalili et~al.}]{hadi2023survey}
Muhammad~Usman Hadi, Rizwan Qureshi, Abbas Shah, Muhammad Irfan, Anas Zafar, Muhammad~Bilal Shaikh, Naveed Akhtar, Jia Wu, Seyedali Mirjalili, et~al. 2023.
\newblock A survey on large language models: Applications, challenges, limitations, and practical usage.
\newblock \emph{Authorea Preprints}.

\bibitem[{Hu et~al.(2022)Hu, Shen, Wallis, Allen-Zhu, Li, Wang, Wang, and Chen}]{hu2022lora}
Edward~J Hu, Yelong Shen, Phillip Wallis, Zeyuan Allen-Zhu, Yuanzhi Li, Shean Wang, Lu~Wang, and Weizhu Chen. 2022.
\newblock \href {https://openreview.net/forum?id=nZeVKeeFYf9} {Lo{RA}: Low-rank adaptation of large language models}.
\newblock In \emph{International Conference on Learning Representations}.

\bibitem[{Hu et~al.(2023)Hu, Tang, Zuo, Wang, Song, Lou, Jiao, and Charles}]{hu2023evoke}
Xinyu Hu, Pengfei Tang, Simiao Zuo, Zihan Wang, Bowen Song, Qiang Lou, Jian Jiao, and Denis Charles. 2023.
\newblock Evoke: Evoking critical thinking abilities in llms via reviewer-author prompt editing.
\newblock \emph{ICLR 2024}.

\bibitem[{Hunt(2003)}]{hunt2003concept}
Darwin~P Hunt. 2003.
\newblock The concept of knowledge and how to measure it.
\newblock \emph{Journal of intellectual capital}, 4(1):100--113.

\bibitem[{Liang et~al.(2023)Liang, He, Jiao, Wang, Wang, Wang, Yang, Tu, and Shi}]{liang2023encouraging}
Tian Liang, Zhiwei He, Wenxiang Jiao, Xing Wang, Yan Wang, Rui Wang, Yujiu Yang, Zhaopeng Tu, and Shuming Shi. 2023.
\newblock Encouraging divergent thinking in large language models through multi-agent debate.
\newblock \emph{arXiv preprint arXiv:2305.19118}.

\bibitem[{Lu et~al.(2023)Lu, Schuff, and Gurevych}]{lu2023prompts}
Sheng Lu, Hendrik Schuff, and Iryna Gurevych. 2023.
\newblock How are prompts different in terms of sensitivity?
\newblock \emph{arXiv preprint arXiv:2311.07230}.

\bibitem[{Phuong and Lampert(2019)}]{phuong2019towards}
Mary Phuong and Christoph Lampert. 2019.
\newblock Towards understanding knowledge distillation.
\newblock In \emph{International conference on machine learning}, pages 5142--5151. PMLR.

\bibitem[{Pryzant et~al.(2023{\natexlab{a}})Pryzant, Iter, Li, Lee, Zhu, and Zeng}]{pryzant2023automatic_gradient}
Reid Pryzant, Dan Iter, Jerry Li, Yin~Tat Lee, Chenguang Zhu, and Michael Zeng. 2023{\natexlab{a}}.
\newblock Automatic prompt optimization with" gradient descent" and beam search.
\newblock \emph{arXiv preprint arXiv:2305.03495}.

\bibitem[{Pryzant et~al.(2023{\natexlab{b}})Pryzant, Iter, Li, Lee, Zhu, and Zeng}]{pryzant2023automatic}
Reid Pryzant, Dan Iter, Jerry Li, Yin~Tat Lee, Chenguang Zhu, and Michael Zeng. 2023{\natexlab{b}}.
\newblock Automatic prompt optimization with" gradient descent" and beam search.
\newblock \emph{arXiv preprint arXiv:2305.03495}.

\bibitem[{Roy and Roth(2015)}]{roy-roth-2015-solving}
Subhro Roy and Dan Roth. 2015.
\newblock \href {https://doi.org/10.18653/v1/D15-1202} {Solving general arithmetic word problems}.
\newblock In \emph{Proceedings of the 2015 Conference on Empirical Methods in Natural Language Processing}, pages 1743--1752, Lisbon, Portugal. Association for Computational Linguistics.

\bibitem[{Rubin et~al.(2021)Rubin, Herzig, and Berant}]{rubin2021learning}
Ohad Rubin, Jonathan Herzig, and Jonathan Berant. 2021.
\newblock Learning to retrieve prompts for in-context learning.
\newblock \emph{arXiv preprint arXiv:2112.08633}.

\bibitem[{Scerbo et~al.(2019)Scerbo, Calhoun, and Hui}]{scerbo2019research}
Mark~W Scerbo, Aaron~W Calhoun, and Joshua Hui. 2019.
\newblock Research and hypothesis testing: Moving from theory to experiment.
\newblock \emph{Healthcare Simulation Research: A Practical Guide}, pages 161--167.

\bibitem[{Su et~al.(2022)Su, Wang, Qin, Chan, Lin, Wang, Wen, Liu, Li, Li, Hou, Sun, and Zhou}]{su-etal-2022-transferability}
Yusheng Su, Xiaozhi Wang, Yujia Qin, Chi-Min Chan, Yankai Lin, Huadong Wang, Kaiyue Wen, Zhiyuan Liu, Peng Li, Juanzi Li, Lei Hou, Maosong Sun, and Jie Zhou. 2022.
\newblock \href {https://doi.org/10.18653/v1/2022.naacl-main.290} {On transferability of prompt tuning for natural language processing}.
\newblock In \emph{Proceedings of the 2022 Conference of the North American Chapter of the Association for Computational Linguistics: Human Language Technologies}, pages 3949--3969, Seattle, United States. Association for Computational Linguistics.

\bibitem[{Wang et~al.(2023)Wang, Li, Wang, Bai, Luo, Zhang, Jojic, Xing, and Hu}]{wang2023promptagent}
Xinyuan Wang, Chenxi Li, Zhen Wang, Fan Bai, Haotian Luo, Jiayou Zhang, Nebojsa Jojic, Eric~P Xing, and Zhiting Hu. 2023.
\newblock Promptagent: Strategic planning with language models enables expert-level prompt optimization.
\newblock \emph{ICLR 20247}.

\bibitem[{Wei et~al.(2022)Wei, Wang, Schuurmans, Bosma, Xia, Chi, Le, Zhou et~al.}]{wei2022chain}
Jason Wei, Xuezhi Wang, Dale Schuurmans, Maarten Bosma, Fei Xia, Ed~Chi, Quoc~V Le, Denny Zhou, et~al. 2022.
\newblock Chain-of-thought prompting elicits reasoning in large language models.
\newblock \emph{Advances in neural information processing systems}, 35:24824--24837.

\bibitem[{Wen et~al.(2024)Wen, Jain, Kirchenbauer, Goldblum, Geiping, and Goldstein}]{wen2024hard}
Yuxin Wen, Neel Jain, John Kirchenbauer, Micah Goldblum, Jonas Geiping, and Tom Goldstein. 2024.
\newblock Hard prompts made easy: Gradient-based discrete optimization for prompt tuning and discovery.
\newblock \emph{Advances in Neural Information Processing Systems}, 36.

\bibitem[{Xia et~al.(2023)Xia, Zheng, Li, Zhuang, Zhou, Qiu, Li, Lin, and Song}]{xia2023flash}
Haojun Xia, Zhen Zheng, Yuchao Li, Donglin Zhuang, Zhongzhu Zhou, Xiafei Qiu, Yong Li, Wei Lin, and Shuaiwen~Leon Song. 2023.
\newblock Flash-llm: Enabling cost-effective and highly-efficient large generative model inference with unstructured sparsity.
\newblock \emph{arXiv preprint arXiv:2309.10285}.

\bibitem[{Xie et~al.(2021)Xie, Raghunathan, Liang, and Ma}]{xie2021explanation}
Sang~Michael Xie, Aditi Raghunathan, Percy Liang, and Tengyu Ma. 2021.
\newblock An explanation of in-context learning as implicit bayesian inference.
\newblock \emph{arXiv preprint arXiv:2111.02080}.

\bibitem[{Yang et~al.(2023)Yang, Li, and Liu}]{yang2023failures}
Zeyuan Yang, Peng Li, and Yang Liu. 2023.
\newblock Failures pave the way: Enhancing large language models through tuning-free rule accumulation.
\newblock \emph{arXiv preprint arXiv:2310.15746}.

\bibitem[{Ye et~al.(2023)Ye, Axmed, Pryzant, and Khani}]{ye2023prompt}
Qinyuan Ye, Maxamed Axmed, Reid Pryzant, and Fereshte Khani. 2023.
\newblock Prompt engineering a prompt engineer.
\newblock \emph{arXiv preprint arXiv:2311.05661}.

\bibitem[{Zhang et~al.(2024)Zhang, Madaan, Gao, Zheng, Mishra, Yang, Tandon, and Alon}]{zhang2024context}
Tianjun Zhang, Aman Madaan, Luyu Gao, Steven Zheng, Swaroop Mishra, Yiming Yang, Niket Tandon, and Uri Alon. 2024.
\newblock In-context principle learning from mistakes.
\newblock \emph{arXiv preprint arXiv:2402.05403}.

\bibitem[{Zhong et~al.(2024)Zhong, Ding, Liu, Du, and Tao}]{zhong2024panda}
Qihuang Zhong, Liang Ding, Juhua Liu, Bo~Du, and Dacheng Tao. 2024.
\newblock Panda: Prompt transfer meets knowledge distillation for efficient model adaptation.
\newblock \emph{IEEE Transactions on Knowledge and Data Engineering}.

\bibitem[{Zhou et~al.(2022{\natexlab{a}})Zhou, Muresanu, Han, Paster, Pitis, Chan, and Ba}]{zhou2022large_ape}
Yongchao Zhou, Andrei~Ioan Muresanu, Ziwen Han, Keiran Paster, Silviu Pitis, Harris Chan, and Jimmy Ba. 2022{\natexlab{a}}.
\newblock Large language models are human-level prompt engineers.
\newblock \emph{arXiv preprint arXiv:2211.01910}.

\bibitem[{Zhou et~al.(2022{\natexlab{b}})Zhou, Muresanu, Han, Paster, Pitis, Chan, and Ba}]{zhou2022large}
Yongchao Zhou, Andrei~Ioan Muresanu, Ziwen Han, Keiran Paster, Silviu Pitis, Harris Chan, and Jimmy Ba. 2022{\natexlab{b}}.
\newblock Large language models are human-level prompt engineers.
\newblock \emph{arXiv preprint arXiv:2211.01910}.

\bibitem[{Zhu et~al.(2023)Zhu, Xue, Chen, Zhou, Tang, Schuurmans, and Dai}]{zhu2023large}
Zhaocheng Zhu, Yuan Xue, Xinyun Chen, Denny Zhou, Jian Tang, Dale Schuurmans, and Hanjun Dai. 2023.
\newblock Large language models can learn rules.
\newblock \emph{arXiv preprint arXiv:2310.07064}.

\end{thebibliography}
